\pdfoutput=1

\documentclass[11pt]{article}

\usepackage{naacl2021}

\usepackage{times}
\usepackage{latexsym}
\usepackage{setspace}
\usepackage{textcomp}
\usepackage[T1]{fontenc}

\usepackage[utf8]{inputenc}
\usepackage{colortbl}

\usepackage{microtype}

\usepackage{multirow}
\usepackage{graphicx}
\usepackage{algorithm}
\usepackage{algorithmic}
\usepackage[mathscr]{euscript}
\usepackage{subcaption}
\usepackage{times}
\usepackage{latexsym}
\usepackage{tabularx}
\usepackage{multirow}
\usepackage{booktabs}
\usepackage{array}
\usepackage{footnote}
\usepackage{graphicx}
\usepackage{dingbat}
\usepackage{svg}
\usepackage{amsmath}
\usepackage[normalem]{ulem}
\usepackage{bchart}
\usepackage{pgfplots}
\usepackage{hyperref}
\usepackage{array}
\usepackage{amssymb}
\usepackage{pifont}
\usepackage{todonotes}
\usepackage{adjustbox}
\usepackage{xr}
\usepackage{collcell}

\usepackage{siunitx}
\usepackage[inline]{enumitem}
\newcommand{\cmmnt}[1]{}

\makesavenoteenv{tabular}
\makesavenoteenv{table*}
\newcolumntype{L}[1]{%
  >{\raggedright\let\newline\\\arraybackslash\hspace{0pt}}p{#1}%
}
\usetikzlibrary{shapes.geometric}

\usepackage{pgfplotstable}

\pgfplotstableset{
    color cells/.style={
        col sep=comma,
        string type,
        postproc cell content/.code={%
                \pgfkeysalso{@cell content=\rule{0cm}{2.4ex}\cellcolor{green!##10}{#1}\pgfmathtruncatemacro\number{##1}\ifnum\number<0\cellcolor{red!10}\fi##1}%
                },
        columns/x/.style={
            column name={},
            postproc cell content/.code={}
        }
    }
}

\definecolor{bblue}{HTML}{4F81BD}
\definecolor{rred}{HTML}{C0504D}
\definecolor{ggreen}{HTML}{9BBB59}
\definecolor{ppurple}{HTML}{9F4C7C}

\newcommand{\setset}{\mathcal{D}}
\newcommand{\metatraindata}{\setset_{\mbox{\tiny{meta-train}}}}
\newcommand{\dev}{\mbox{$Dev$}}
\newcommand{\train}{\mbox{$Train$}}
\newcommand{\metaadaptdata}{\setset_{\mbox{\tiny{meta-adapt}}}}

\newcommand*{\MinNumber}{0}%
\newcommand*{\MaxNumber}{1}

\newcommand{\ApplyGradient}[1]{%
        \pgfmathsetmacro{\PercentColor}{100.0*(#1-\MinNumber)/(\MaxNumber-\MinNumber)}
       \hspace{-0.33em}\colorbox{blue!\PercentColor!black}{}
}

\newcolumntype{B}{>{\collectcell\ApplyGradient}c<{\endcollectcell}}

\usepackage{amsfonts}
\usepackage{booktabs}
\usepackage{siunitx}

\newcommand{\cca}[1]{%
  \ifnum#1<0\cellcolor{red!10}{#1}\else\cellcolor{blue!#1}{#1}\fi
}

\title{X-METRA-ADA: Cross-lingual Meta-Transfer Learning Adaptation to Natural Language Understanding and Question Answering}

\author{
Meryem M\textquotesingle hamdi$^{1}$\thanks{\hspace{1.5mm}Work was started while the first author was a research intern at Adobe.}\hspace{2mm}, Doo Soon Kim$^{2}$, Franck Dernoncourt$^{2}$,\\ \textbf{Trung Bui$^{2}$, Xiang Ren$^{1}$ and Jonathan May}$^{1}$\\
$^{1}$Information Sciences Institute, University of Southern California\\
{\{\tt meryem}, {\tt xiangren}, {\tt jonmay}\}@isi.edu \\
$^{2}$Adobe Research \\
{\{\tt dkim}, {\tt dernonco}, {\tt bui}\}@adobe.com \\
\\}

\begin{document}
\maketitle
\begin{abstract}
Multilingual models, such as M-BERT and XLM-R, have gained increasing popularity, due to their zero-shot cross-lingual transfer learning capabilities. However, their generalization ability is still inconsistent for typologically diverse languages and across different benchmarks. Recently, meta-learning has garnered attention as a promising technique for enhancing transfer learning under low-resource scenarios: particularly for cross-lingual transfer in Natural Language Understanding (NLU).

In this work, we propose \textbf{X-METRA-ADA}, a \textbf{cross}-lingual \textbf{ME}ta-\textbf{TRA}nsfer learning \textbf{ADA}ptation approach for NLU. Our approach adapts MAML, an optimization-based meta-learning approach, to learn to adapt to new languages.
We extensively evaluate our framework on two challenging cross-lingual NLU tasks: multilingual task-oriented dialog and typologically diverse question answering. We show that our approach outperforms naive fine-tuning, reaching competitive performance on both tasks for most languages. Our analysis reveals that X-METRA-ADA can leverage limited data for faster adaptation.
\end{abstract}

\begin{figure}[t!]
\includegraphics[width=0.52\textwidth]{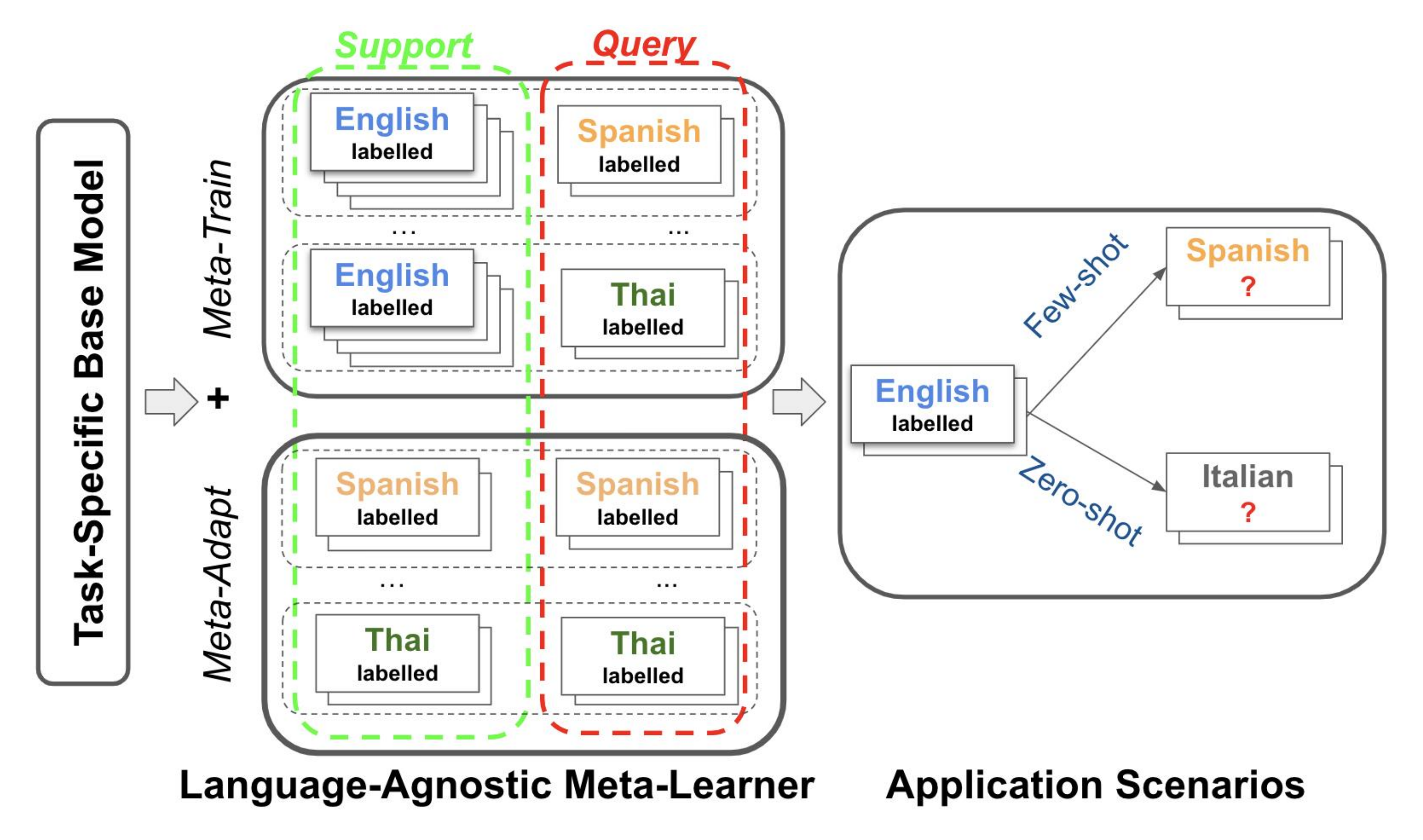}
\caption{\label{meta-learning} An overview of the X-METRA-ADA framework: we use English as the source and Spanish as the target language. The \underline{meta-train} stage transfers from the source to the target languages, while the \underline{meta-adaptation} further adapts the model to the target language. The application is \textbf{few-shot} if the test language is seen in any stage of X-METRA-ADA; or \textbf{zero-shot} if the test language is unseen.}
\end{figure}
\vspace{-0.5cm}

\section{Introduction} 

Cross-lingual transfer learning is a technique used to adapt a model trained on a downstream task in a source language to directly generalize to the task in new languages. It aims to come up with common cross-lingual representations and leverages them to bridge the divide between resources to make any NLP application scale to multiple languages. This is particularly useful for data-scarce scenarios, as it reduces the need for API calls implied by machine translation or costly task-specific annotation for new languages. 

Transformer-based contextualized embeddings and their multilingual counterparts such as  M-BERT~\cite{bert-devlin-naacl19} have become popular as off-the-shelf representations for cross-lingual transfer learning. While these multilingual representations exhibit some cross-lingual capability even for languages with low lexical overlap with English, the transfer quality is reduced for languages that exhibit different typological characteristics~\cite{multibert-pires-acl19}.

The generalization of such representations has been extensively evaluated on traditional tasks such as Part-of-Speech (POS) tagging, Named Entity Recognition (NER) and Cross-lingual Document Classification (CLDC) (\citealp{deppars-ahmad-emnlp19}; ~\citealp{betobentz-wu-emnlp19}; ~\citealp{zeroner-saiful-aaai20}; ~\citealp{mldoc-schwenk-lrec18}), with ever-growing open community annotation efforts like Universal Dependencies~\cite{ud2-nivre-lrec20} and CoNLL shared tasks (\citealp{conll02-sang-coling02}; ~\citealp{conll03-sang-naacl03}). On the other hand, cross-lingual Natural Language Understanding (NLU) tasks have gained less attention, with smaller benchmark datasets that cover a handful of languages and don't truly model linguistic variety (\citealp{xnli-conneau-emnlp18}; ~\citealp{xsquad-artetxe-acl20}). Natural Language Understanding tasks are critical for dialog systems, as they make up an integral part of the dialog pipeline. Understanding and improving the mechanism behind cross-lingual transfer for natural language understanding in dialog systems require evaluations on more challenging and typologically diverse benchmarks.

Numerous approaches have attempted to build stronger cross-lingual representations on top of those multilingual models; however, most require parallel corpora (\citealp{clbt-wang-emnlp19}; ~\citealp{xlm-lample-neurips19}) and are biased towards high-resource and balanced setups. This fuels the need for a method that doesn't require explicit cross-lingual alignment for faster adaptation to low-resource setups. 

Meta-learning, a method for ``learning to learn'', has found favor especially among the computer vision and speech recognition communities (\citealp{nichol-arxiv-18};~\citealp{meta-data-20};~\citealp{winata-meta-learn-acl20}). Meta-learning has been used for machine translation~\cite{metamt-gu-emnlp18}, few-shot relation classification~\cite{fewrel2-gao-emnlp19}, and on a variety of GLUE tasks~\cite{dou-etal-2019-investigating}. Recently, \citet{xlingualnlumaml-nooralahzadeh-arxiv20} apply the MAML~\cite{maml-finn-icml17} algorithm to cross-lingual transfer learning for XNLI~\cite{xnli-conneau-emnlp18} and MLQA~\cite{lewis-mlqa-acl20}, NLU tasks that are naturally biased towards machine translation-based solutions. \citeauthor{xlingualnlumaml-nooralahzadeh-arxiv20} are able to show improvement over strong multilingual models, including M-BERT. However, they mainly show the effects of meta-learning as a first step in a framework that relies on supervised fine-tuning, making it difficult to properly compare and contrast both approaches.

We study cross-lingual meta-transfer learning from a different perspective. We distinguish between meta-learning and fine-tuning and design systematic experiments to analyze the added value of meta-learning compared to naive fine-tuning. We also build our analysis in terms of more typologically diverse cross-lingual NLU tasks: Multilingual Task-Oriented Dialogue System (MTOD)~\cite{xlingualnlu-schuster-naacl19} and Typologically Diverse Question Answering (TyDiQA)~\cite{tydiqa-clark-tacl20}. While XNLI is a classification task, MTOD is a joint classification and sequence labelling task and is more typologically diverse. TyDiQA is not a classification task, but we show how meta-learning can be applied usefully to it. We also show greater performance improvements from meta-learning than fine-tuning on transfer between typologically diverse languages.

To the best of our knowledge, we are the first to conduct an extensive analysis applied to MTOD and TyDiQA to evaluate the quality of cross-lingual meta-transfer. Our contributions are three-fold:
\vspace{-0.3cm}
\begin{itemize}[leftmargin=*]
\itemsep0em 
\item {Proposing X-METRA-ADA,\footnote{We release our code at: \url{github.com/meryemmhamdi1/meta_cross_nlu_qa}.} a language-agnostic meta-learning framework (Figure~\ref{meta-learning}), and extensively evaluating it.}
\vspace{-0.1cm}
\item {Applying X-METRA-ADA to two challenging cross-lingual and typologically diverse task-oriented dialog and QA tasks, which includes recipes for constructing appropriate meta-tasks (Section~\ref{sec:meta-learn-data}).}
\vspace{-0.1cm}
\item {Analyzing the importance of different components in cross-lingual transfer and the scalability of our approach across different k-shot and down-sampling configurations (Section~\ref{ana-results}).}
\end{itemize}

\section{Methodology}
\label{sec:methodology}

We make use of optimization-based meta-learning on top of pre-trained models with two levels of adaptation to reduce the risk of over-fitting to the target language: (i) \textbf{meta-training} from the source language to the target language(s) (ii) \textbf{meta-adaptation} on the same target language(s) for more language-specific adaptation (Figure~\ref{meta-learning}). 

We apply our approach to two cross-lingual downstream tasks: MTOD (Section~\ref{sec:nlu}) and TyDiQA (Section~\ref{sec:qa}). We start by describing the base architectures for both tasks, before explaining how they are incorporated into our meta-learning pipeline. Applying meta-learning to a task requires the construction of multiple `pseudo-tasks', which are instantiated as pairs of datasets. We describe this construction for our downstream tasks in Section~\ref{sec:meta-learn-data}. Finally, we present our X-METRA-ADA algorithm (Section~\ref{sec:meta-learn-data-app}).

\subsection{Multilingual Task-Oriented Dialog (MTOD)}
\label{sec:nlu}

\begin{figure}[t!]
\centering
\includegraphics[width=0.35\textwidth]{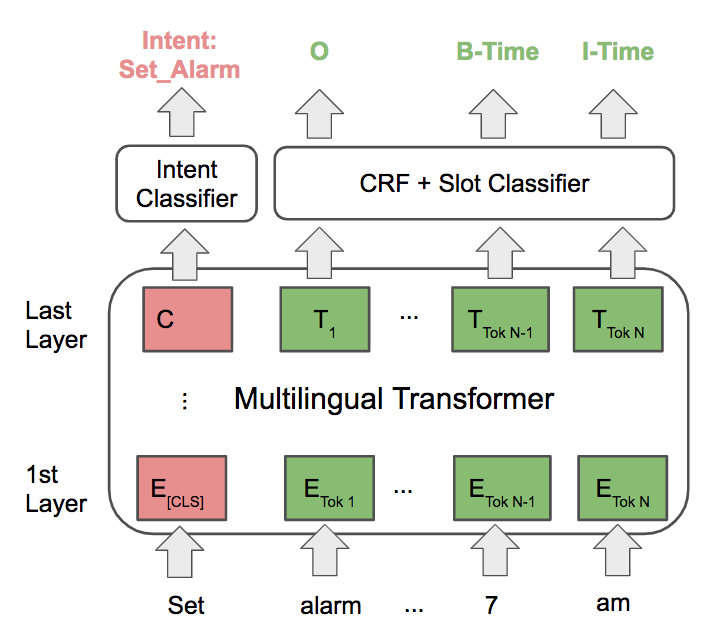}
\caption{\label{mtod-base-model} Architecture of Base MTOD.}
\end{figure}

Similar to the architecture in~\citet{giuseppe-jointBERT-19}, we model MTOD's intent classification and slot filling subtasks jointly. For that purpose, we use a joint text classification and sequence labeling framework with feature representation based on Transformer \cite{vaswani-transformers-nips17}. More specifically, given a multilingual pre-trained model, we use it to initialize the word-piece embeddings layer. Then, we add on top of it a text classifier to predict the intent from the $[CLS]$ token representation and a sequence labeling layer in the form of a linear layer to predict the slot spans (in BIO annotation), as shown in Figure~\ref{mtod-base-model}. We optimize parameters using the sum of both intent and CRF based slot losses.

\subsection{Typologically Diverse Question Answering (TyDiQA)}
\label{sec:qa}

\begin{figure}[h!]
  \centering
  \includegraphics[width=0.4\textwidth]{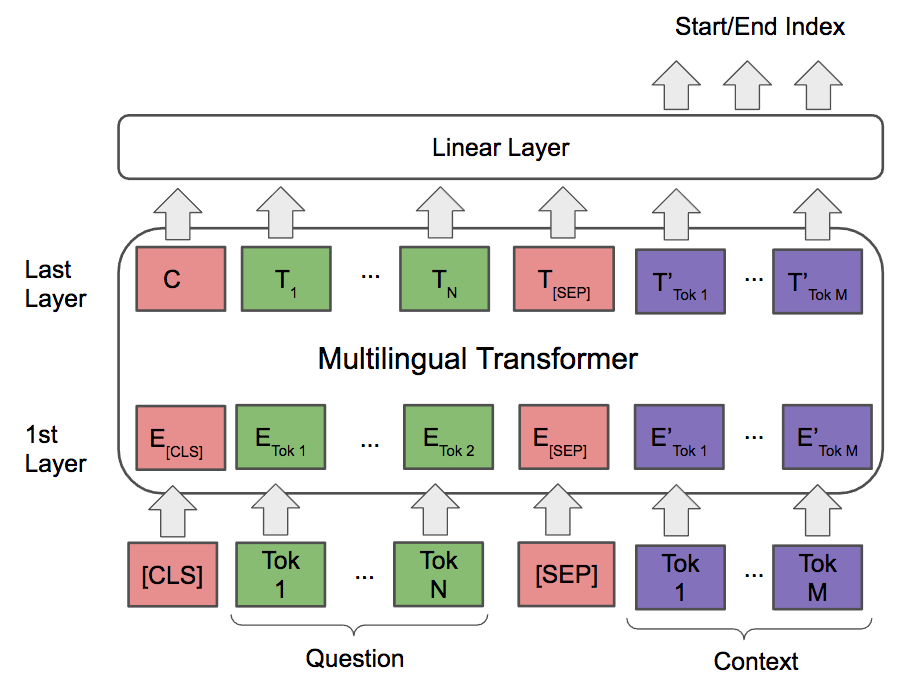}
  \caption{\label{fig:qa-base}Question Answering Base Model.}
\end{figure}

Inspired by~\citet{xtreme-hu-arxiv20}, we apply to TyDiQA the same architecture as the original BERT fine-tuning procedure for question answering on SQuAD~\cite{bert-devlin-naacl19}. Specifically, the input question (after prepending it with a $[CLS]$ token) and the context are concatenated as a single packed sequence separated by a $[SEP]$ token. Then, the embeddings of the context are fed to a linear layer plus a softmax to compute the probability that each token is the START or END of the answer. The whole architecture is fine-tuned by optimizing for the joint loss over the START and END predictions. Any START and END positions that are outside of the scope of the context end up being truncated because of Transformer-based embeddings length limitations and are ignored during training. Figure~\ref{fig:qa-base} illustrates the architecture.

\subsection{Psuedo-task Datasets}
\label{sec:meta-learn-data}

Meta-learning is distinguished from fine-tuning in that the former seeks an initialization point that is maximally useful to multiple downstream learning tasks, while the latter seeks to directly optimize a downstream `child' task from the initialization point of a `parent' task.  To apply meta-learning to data scenarios that more closely fit fine-tuning, we construct multiple `pseudo-tasks' by subsampling from parent and child task datasets. A pseudo-task is defined as a tuple $T = (S, Q)$, where each of $S$ and $Q$ are labeled samples. In the inner loops of meta-learning, the loss on $Q$ from a model trained on $S$ is used to adapt the initialization point (where $Q$ and $S$ are referred to as the \textit{query} and \textit{support} in meta-learning literature). Pseudo-tasks are constructed in such a way as to make them balanced and non-overlapping.  We describe our approach for each task below.

\subsubsection{MTOD Pseudo-task Construction}
\label{meta-mtod-tasks}

MTOD labeled data consists of a sentence from a dialogue along with a sentence-level intent label and subsequence slot labels. From the available data, we draw a number of task sets $\mathcal{T}$; each $T=(S, Q) \in \mathcal{T}$ consists of $k$ intent and slot-labeled items per intent class in $S$ and $q$ items per class in $Q$. Although carefully arranged to have the same number of items per class per task in each of the support and the query sets, the same task splits are used for slot prediction as well. During meta-training and meta-adaptation, task batches are sampled randomly from $\mathcal{T}$.

\subsubsection{QA Pseudo-task Construction}
\label{meta-qa-tasks}

Unlike MTOD, QA is not a standard classification task with fixed classes; thus, it is not directly amenable to class distribution balancing across pseudo-task query and support sets. To construct pseudo-tasks for QA from the available (question, context, answer) span triplet data, we use the following procedure: We draw a task $T = (S, Q)$, by first randomly drawing $q$ triplets, forming $Q$. For each triplet $t$ in $Q$, we draw the $k/q$ \textit{most similar} triplets to $t$ from the remaining available data, thus forming $S$.\footnote{Thus $k$ is constrained to be a multiple of $q$.} For two triplets $t_1$, $t_2$ we define similarity as $\cos(f(t_1), f(t_2))$, where $f(.)$ is a representation of the concatenation of the triplet elements delimited by a space; we use a cross-lingual extension to SBERT's pre-trained model~\cite{reimers-2019-sentence-bert, reimers2020studentteacher}.

\subsubsection{Cross-lingual extension}

In the original MAML~\cite{maml-finn-icml17}, in every iteration we sample a task set $\mathcal{T}$ from a single distribution $\setset$, and the support and query sets in a single task $T$ would be drawn from a common space. We distinguish between the distributions $\metatraindata$ and $\metaadaptdata$, which correspond to the two levels of adaptation introduced in Section~\ref{sec:methodology} and explained below in Section~\ref{sec:meta-learn-data-app}. 

To enable cross-lingual transfer, we draw data for the support set of tasks in $\metatraindata$ from task data in the high-resource base language (English, in our experiments). For the query set in $\metatraindata$ and for both support and query sets in $\metaadaptdata$, we sample from task data in the language to be evaluated. 

\subsection{X-METRA-ADA Algorithm}
\label{sec:meta-learn-data-app}
Following the notation described in the above sections, we present our algorithm X-METRA-ADA, our adaptation  of MAML to cross-lingual transfer learning in two stages. In each stage we use the procedure outlined in Algorithm~\ref{algo-1}. We start by sampling a batch of tasks from distribution $\setset$. For every task $T_{j}=(S_{j}, Q_{j})$, we update $\theta_{j}$ over $n$ steps using batches drawn from $S_{j}$. At the end of this inner loop, we compute the gradients with respect to the loss of $\theta_j$ on $Q_{j}$. At the end of all tasks of each batch, we sum over all pre-computed gradients and update $\theta$, thus completing one outer loop. The difference between meta-train and meta-adapt stages comes down to the parameters and hyperparameters passed into Algorithm~\ref{algo-1}. 

\begin{itemize}[leftmargin=*]
\vspace{-0.2cm}
    \item \textbf{Meta-train}: This stage is similar to classical MAML. Task sets are sampled from $\metatraindata$, which uses high-resource (typically English) data in support sets and low-resource data in the query sets. The input model $\theta_B$ is typically a pre-trained multilingual downstream base model, and we use hyperparameters $n=5, \alpha=\num{1e-3}$ and $\beta=\num{1e-2}$ for MTOD and $\alpha=\beta=\num{3e-5}$ for QA. 
    \item \textbf{Meta-adapt}: During this stage, we ensure the model knows how to learn from examples within the target language under a low-resource regime. Task sets are sampled from $\metaadaptdata$, which uses low-resource data in both support and query sets. The input model is the optimization resulting from meta-train, and we use hyperparameters $n=5, \alpha=\num{1e-3}$ and $\beta=\num{1e-2}$ for MTOD and $\alpha=\beta=\num{3e-5}$ for QA.
\end{itemize}

\begin{algorithm}[t!]
\caption{X-METRA-ADA
}
\begin{small}
\begin{algorithmic}[1]
\REQUIRE Task set distribution $\setset$, pre-trained learner $B$ with parameters $\theta_{B}$, meta-learner $M$ with parameters ($\theta$, $\alpha$, $\beta$, $n$)
\STATE Initialize $\theta \leftarrow \theta_{B}$ 
\WHILE{not done}
\STATE Sample batch of tasks $\mathcal{T}= \{T_1, T_2, \ldots T_b\} \sim \setset$
\FOR{all $T_j = (S_j, Q_j)$ in $\mathcal{T}$}
\STATE Initialize $\theta_{j} \leftarrow \theta$ 
\FOR{$t=1\ldots n$}
\STATE Evaluate $\partial B_{\theta_{j}} / \partial \theta_{j}  = \nabla_{\theta_{j}} \mathcal{L}^{S_{j}}_{T_{j}}(B_{\theta_{j}})$
\STATE Update $\theta_{j} = \theta_{j} - \alpha \partial B_{\theta_{j}} / \partial \theta_{j} $

\ENDFOR

\STATE Evaluate query loss $\mathcal{L}^{Q_{j}}_{T_j}(B_{\theta_{j}}) $ and save it for outer loop
\ENDFOR

\STATE Update $\theta \leftarrow \theta - \beta \nabla_{\theta} \sum^{b}_{j=1} \mathcal{L}^{Q_{j}}_{T_j}(B_{\theta_{j}}) $ 

\ENDWHILE
\end{algorithmic}
\end{small}
\label{algo-1}
\end{algorithm}

\vspace{-0.2cm}

\section{Experimental Setup}

\subsection{Datasets}
For dialogue intent prediction, we use the Multilingual Task-Oriented Dialogue (MTOD)~\citep{xlingualnlu-schuster-naacl19} dataset.  MTOD covers 3 languages (English, Spanish, and Thai), 3 domains (alarm, reminder, and weather), 12 intent types, and 11 slot types.\footnote{We follow the same pre-processing and evaluation as \citet{mixedNLU-zihan-aaai20}.} We train models with the English training data ($\train$) but for the other languages we use the provided development sets ($\dev$) to further our goals to analyze methods of few-shot transfer. We evaluate on the provided test sets. Moreover, we evaluate on an in-house dataset of 7 languages.\footnote{More details are included in the Appendix \ref{app-res-inhouse}.} 

For QA, we use the Typologically Diverse QA (TyDiQA-GoldP)~\cite{tydiqa-clark-tacl20} dataset. TyDiQA is a typologically diverse question answering dataset covering 11 languages. Like~\citet{xtreme-hu-arxiv20}, we use a simplified version of the primary task. Specifically, we discard questions that don't have an answer and use only the gold passage as context, keeping only the short answer and its spans. This makes the task similar to XQuAD and MLQA, although unlike these tasks, the questions are written without looking at the answers and without machine translation. As with MTOD, we use the English training data as $\train$. Since development sets are not specified for MTOD, we instead  reserve 10\% of the training data in each of the other languages as $\dev$. We report on the provided test sets. Statistics of datasets for both tasks can be found in Appendix~\ref{app-data-stats}.

\subsection{Evaluation}
\label{eval-categ}
In order to fairly and consistently evaluate our approach to few-shot transfer learning via meta-learning and to  ablate components of the method, we design a series of experiments based on both internal and external baselines. Our internal baselines ablate the effect of the X-METRA-ADA algorithm vs. conventional fine-tuning from a model trained on a high-resource language by keeping the data sets used for training constant. As our specific data conditions are not reproduced in any externally reported results on these tasks, we instead compare to other reported results using English-only or entirely zero-shot training data. 
\paragraph{Internal Evaluation}
We design the following fine-tuning/few-shot schemes:
\begin{itemize}[leftmargin=*]
    \vspace{-0.2cm}
    \item {\textit{PRE}: An initial model is fine-tuned on the $\train$ split of English only and then evaluated on new languages with no further tuning or adaptation. This strawman baseline has exposure to English task data only.}
    \vspace{-0.2cm}
    \item{\textit{MONO}: An initial model is fine-tuned on the $\dev$ split of the target language. This baseline serves as a comparison for standard fine-tuning (FT, below), which shows the value of combining MONO and PRE.}
    \vspace{-0.2cm}
    \item{ \textit{FT}: We fine-tune the PRE model on the $\dev$ split of the target language. This is a standard transfer learning approach that combines PRE and MONO. }
     \vspace{-0.2cm}
    \item{ \textit{FT w/EN}: Like FT, except both the $\dev$ split of the target language and the $\train$ split of English are used for fine-tuning. This is used for dataset equivalence with X-METRA-ADA (below).}
    \vspace{-0.2cm}
    \item{\textit{X-METRA}: We use the PRE model as $\theta_B$ for meta-train, the $Train$ split from English to form support sets in $D_{meta-train}$, and all of the $Dev$ split of the target language to form query sets in $D_{meta-train}$.}
    \vspace{-0.2cm}
    \item{\textit{X-METRA-ADA}: We use the PRE model as $\theta_B$ for meta-train, the $\train$ split from English to form support sets in $\metatraindata$. For MTOD, we use 75\% of the $\dev$ split of the target language to form query sets in $\metatraindata$. We use the remaining 25\% of the $\dev$ split of the target language for both the support and query sets of $\metaadaptdata$. For QA, we use ratios of 60\% for $\metatraindata$  and 40\% for $\metaadaptdata$.}
\end{itemize}
All models are ultimately fine-tuned versions of BERT and all have access to the same task training data relevant for their variant. That is, X-METRA-ADA and PRE both see the same English $\train$ data and MONO, FT, and X-METRA-ADA see the same target language $\dev$ data. However, since X-METRA-ADA uses both $\train$ and $\dev$ to improve upon PRE, and FT only uses $\dev$, we make an apples-to-apples comparison, data-wise, by including FT w/EN experiments as well.

\paragraph{External Baselines}
We focus mainly on transfer learning baselines from contextualized embeddings for a coherent external comparison; supervised experiments on target language data such as those reported in \citet{xlingualnlu-schuster-naacl19} are inappropriate for comparison because they use much more in-language labeled data to train. The experiments we compare to are zero-shot in the sense that they are not trained directly on the language-specific task data. However, most of these external baselines involve some strong cross-lingual supervision either through cross-lingual alignment or mixed-language training. We also include machine translation baselines, which are often competitive and hard to beat. Our work, by contrast, uses no parallel language data or resources beyond pretrained multilingual language models, labeled English data, and few-shot labeled target language data. To the best of our knowledge, we are the first to explore cross-lingual meta-transfer learning for those benchmarks, so we only report on our X-METRA-ADA approach in addition to those baselines. 

For MTOD, then, we focus on the following external baselines:
\vspace{-0.1cm}
\begin{itemize}[leftmargin=*]
\itemsep0em 
\vspace{-0.1cm}
\item{\textit{Cross-lingual alignment-based approaches:} We use \underline{MCoVe}, a multilingual version of contextualized word vectors with an autoencoder objective as reported by~\citet{xlingualnlu-schuster-naacl19} in addition to \underline{M-BERT} \cite{mixedNLU-zihan-aaai20}. We also include  \underline{XLM} trained on Translation Language Modeling (TLM) + Masked Language Modeling (MLM) \cite{xlm-lample-neurips19} as enhanced by Transformer and mixed-training as reported by \citet{mixedNLU-zihan-aaai20}.}
\vspace{-0.2cm}
\item{\textit{Mixed-language training approaches}: We use M-BERT + Transformer + mixed training using data from the dialogue domain: from (a) human-based word selection (\underline{MLT$_{H}$}) and (b) attention-based word selection (\underline{MLT$_{A}$}), both are reported by~\citet{mixedNLU-zihan-aaai20}.}
\vspace{-0.2cm}
\item{\textit{Translation-based approaches}: We use the zero-shot version of \underline{MMTE}, the massively multilingual translation encoder by \citet{mmte-aaai-20} fine-tuned on intent classification. We also include \underline{Translate Train (TTrain)} \cite{xlingualnlu-schuster-naacl19}, which translates English training data into target languages to train on them in addition to the target language training data.}
\end{itemize}
For TyDiQA-GoldP, out of the already mentioned baselines, we use M-BERT, XLM, MMTE, and TTrain (which unlike \cite{xlingualnlu-schuster-naacl19} only translates English to the target language to train on it without data augmentation). In addition to that we also include XLM-R as reported by~\citet{xtreme-hu-arxiv20}.
\vspace{-0.3cm}
\subsection{Implementation Details}
We use M-BERT (bert-base-multilingual-cased)\footnote{\url{github.com/huggingface/transformers} version 3.4.0 pre-trained on 104 languages, including all languages evaluated on in this paper.} with 12 layers as initial models for MTOD and TyDiQA-GoldP in our internal evaluation. We use  xlm-r-distilroberta-base-paraphrase-v1\footnote{\url{github.com/UKPLab/sentence-transformers} which uses XLM-R as the base model.} model for computing similarities when constructing the QA meta-dataset (Section \ref{meta-qa-tasks}).

Our implementation of X-METRA-ADA from scratch uses learn2learn~\cite{arnold2020learn2learn} for differentiation and update rules in the inner loop.\footnote{\url{github.com/learnables/learn2learn}.} We use the first-order approximation option in learn2learn for updating the outer loop, also introduced in \citet{maml-finn-icml17}.  
For each model, we run for 3 to 4 different random initializations (for some experiments like PRE for TyDiQA-GoldP we use only 2 seeds respectively) and report the average and standard deviation of the best model for the few-shot language for each run. We use training loss convergence as a criteria for stopping. For the FT and MONO baselines, we don't have the luxury of $\dev$ performance, since those baselines use the $\dev$ dataset for training.\footnote{All experiments are run using Pytorch version 1.6.0, 1 GeForce RTX P8 GPU of 11MB of memory CUDA version 10.1. The runtime depends on the size of the dev data but most MTOD models take around 3 hours to converge and TyDiQA models take a maximum of 10 hours training (including evaluation at checkpoints).} The $\dev$ set is chosen to simulate a low-resource setup. More details on the hyperparameters used can be found in Appendix~\ref{app-hyperparam}.
\section{Results and Discussion}

\subsection{Zero-shot and Few-shot Cross-Lingual NLU and QA}
\label{experiments}

\begin{table}[h]
\centering
\scalebox{0.68}{
\begin{tabular}{l|ll|ll}  \toprule
\multirow{2}{*}{\textbf{Model}}                                                                                         & \multicolumn{2}{c|}{\textbf{Spanish}} & \multicolumn{2}{c}{\textbf{Thai}} \cmmnt{ & \multicolumn{2}{c}{\textbf{Average}}} \\
                                                                                                                      & Intent Acc      & Slot F1      & Intent Acc      & Slot F1 \cmmnt{& Intent Acc      & Slot F1  }    \\ \toprule 
\rowcolor{lightgray} \multicolumn{5}{c}{External Baselines}   \\ \midrule                 
 MCoVe$^{\dagger}$   & 53.9            & 19.3         & 70.7            & 35.6   \cmmnt{& 62.3 & 27.5 }    
 
 \\ M-BERT$^{\ddagger}$ & 73.7  & 51.7 & 28.1 & 10.6 \\
 MLT$_{H}^{\ddagger}$ & 82.9            & \textbf{74.9}         & 53.8            & 26.1  \cmmnt{& 68.4   & 50.5 }  \\  
MLT$_{A}^{\ddagger}$  & 87.9            & 73.9         &  \underline{73.5}            & 27.1 \cmmnt{&  80.7 & \underline{50.5}} \\
XLM$^{\ddagger}$ &  87.5 & 68.5 & 72.6 & 27.9 \\ 
MMTE$^{+}$ &  \textbf{93.6}             &    -       &  89.6        &  -  \cmmnt{&  \textbf{91.6}   &  -} \\
TTrain$^{\ddagger}$ & 85.4 & \underline{72.9} & \textbf{95.9} & 55.4 \\
\midrule
\rowcolor{lightgray} \multicolumn{5}{c}{Zero-shot Learning}   \\ \midrule 
\textbf{PRE}                                                                              & 70.2            &  38.2         &  45.4            & 12.5    \\ \midrule
\rowcolor{lightgray} \multicolumn{5}{c}{Few-shot Learning}  \\ \midrule

\textbf{MONO}                                                                                   &  82.4 $\pm 6.0$          & 43.9 $\pm 1.5$ &   79.1 $\pm 4.7 $         & 54.1 $\pm 3.9$

  \\ \midrule
\textbf{FT}    &  90.7 $\pm 0.3$ &
\textit{\textbf{67.6}} $\pm 1.3$ & 78.9 $\pm 0.2$ & 66.0 $ \pm 2.1 $ \\      
 \textbf{FT w/EN}                      &  \underline{88.7}   $\pm 0.4$     
 &  67.4   $\pm 1.4$       &   73.7 $\pm 0.1$       &  66.0 $\pm 1.6$   \cmmnt{& 86.6 & 57.9} 
  \\  \midrule 
\textbf{X-METRA} & 89.6 $\pm 1.3$ &  63.6 $\pm 0.5$ & 80.2 $\pm 1.2$ & \textbf{70.4} $\pm 1.2$
 \\ 
\textbf{X-METRA-ADA}    &   \textit{\textbf{92.9}}  $\pm 0.6$            & 60.9  $\pm 1.9$   & \textit{\textbf{86.3}}  $\pm 1.7 $         & \underline{69.6}  $\pm 1.9 $   \cmmnt{&   \undeline{90.9 $\pm 0.9$} & \textbf{58.6 $\pm 0.8$}} 

\\
\bottomrule

\end{tabular}
}
\caption{\label{facebook-results-meta} Performance evaluation on MTOD between meta-learning approaches, fine-tuning internal baselines and external baselines. All our internal experiments use $k=q=6$. Zero-shot learning experiments that train only on English are distinguished from few-shot learning, which include a fair internal comparison. Models in bold indicate our own internal models. \textbf{MONO}, \textbf{FT}, \textbf{FT w/EN}, \textbf{X-METRA}, and \textbf{X-METRA-ADA} models include results for each test language when training on that language. \textbf{FT w/EN} trains jointly on English and only the target language. We highlight the best scores in bold and underline the second best for each language and sub-task. The rest are reported from $^{\dagger}$ \cite{xlingualnlu-schuster-naacl19}, $^{\ddagger}$ \cite{mixedNLU-zihan-aaai20}, and $^{+}$ \cite{mmte-aaai-20}.}
\end{table}

\begin{table*}[ht] 
\small
\centering
\scalebox{0.91}{
\begin{tabular}{l|ccccccc} \toprule        \multirow{2}{*}{\textbf{Model}}  &\multicolumn{7}{c} {\textbf{Test on}} \cmmnt{& \multirow{2}{*}{\textbf{Average}}}
\\ 
& Arabic & Bengali & Finnish & Indonesian  &  Russian & Swahili & Telugu \cmmnt{&}  \\ 
\midrule \rowcolor{lightgray} \multicolumn{8}{c}{External Baselines}   \\ \midrule 
M-BERT$^{\dagger}$                                            & 62.2  & 49.3 & 59.7 & 64.8   & 60.0 & 57.5  & 49.6 \cmmnt{& 57.6}  \\ 
XLM$^{\dagger}$ & 59.4 & 27.2 & 58.2 & 62.5 & 49.2 & 39.4 & 15.5 \\
XLM-R$^{\dagger}$ & 67.6  & 64.0  & 70.5  & 77.4 & 67.0  & 66.1  & 70.1 \\  MMTE$^{\dagger}$ & 63.1 & \underline{55.8} & 53.9  & 60.9 & 58.9 & 63.1 & 54.2 \cmmnt{& 58.6} \\
 
 TTrain$^{\dagger}$ & 61.5 & 31.9 &  62.6  & 68.6 &   53.1 & 61.9 & 27.4  \\ \midrule \rowcolor{lightgray} \multicolumn{8}{c}{Zero-shot Learning}  \\
\midrule 
 \textbf{PRE}                                                     & 62.4 $\pm 2.2$ & 32.9 $\pm 1.4$ & 57.7 $\pm 4.4$ & 67.8 $\pm 3.8$ & 58.2 $\pm 3.7$ & 55.5 $\pm 2.9$ & 33.0 $\pm 5.9$ \cmmnt{& 52.5 $\pm 3.5$} \\ 
 \midrule 
 
 \rowcolor{lightgray}  \multicolumn{8}{c}{Few-shot Learning}   \\ \midrule 
\textbf{MONO} &   74.0 $ \pm 1.1 $ & 38.9 $ \pm 0.8 $ & 63.3 $ \pm 1.5 $ & 67.1 $ \pm 1.9 $ & 54.4 $ \pm 1.3 $ & 60.3 $ \pm 1.2 $  & 61.4 $ \pm 1.0 $ \cmmnt{& 59.9 $\pm 1.2$}
 \\ 
 \midrule   
 \textbf{FT}  &   \underline{77.0} $ \pm 0.3 $ & 51.0 $ \pm 2.7 $ & 70.9 $ \pm 0.4 $ & 77.0 $ \pm 0.4 $ & 64.8 $ \pm 0.4 $ & 70.2 $ \pm 1.7 $ & 65.4 $ \pm 0.6 $
 \\ \midrule 
\textbf{X-METRA}  & \textbf{78.5}  $\pm 0.6$  & 53.2 $\pm 0.5$ & \underline{72.7}   $\pm 0.4$ & \textbf{77.7} $\pm 0.2$  &  \underline{66.1} $\pm 0.1$ & \textbf{71.7} $\pm 0.2$ &   \underline{66.6} $\pm 0.4$ \cmmnt{& \textbf{69.5} $\pm 0.3$}  \\
 
\textbf{X-METRA-ADA} & 76.6  $\pm 0.1$  & \textbf{57.8} $\pm 0.6$ & \textbf{73.0} $\pm 0.3$ & \underline{77.3} $\pm 0.1$  &  \textbf{66.9} $\pm 0.1$ & \underline{70.3} $\pm 0.2$ &   \textbf{72.8} $\pm 0.1$  \\
\bottomrule
\end{tabular}
}
\caption{\label{qa-meta-short} F1 comparison on TyDiQA-GoldP between different meta-learning approaches, fine tuning and external baselines. We highlight the best scores in bold and underline the second best for each language. Our own models are in bold, whereas the rest are reported from $^{\dagger}$ \cite{xtreme-hu-arxiv20}. This is using $k= q=6$.}
\end{table*}

\begin{figure*}[h!]
 \centering
 \begin{subfigure}[b]{0.48\textwidth}
 \centering
\includegraphics[width=1\textwidth]{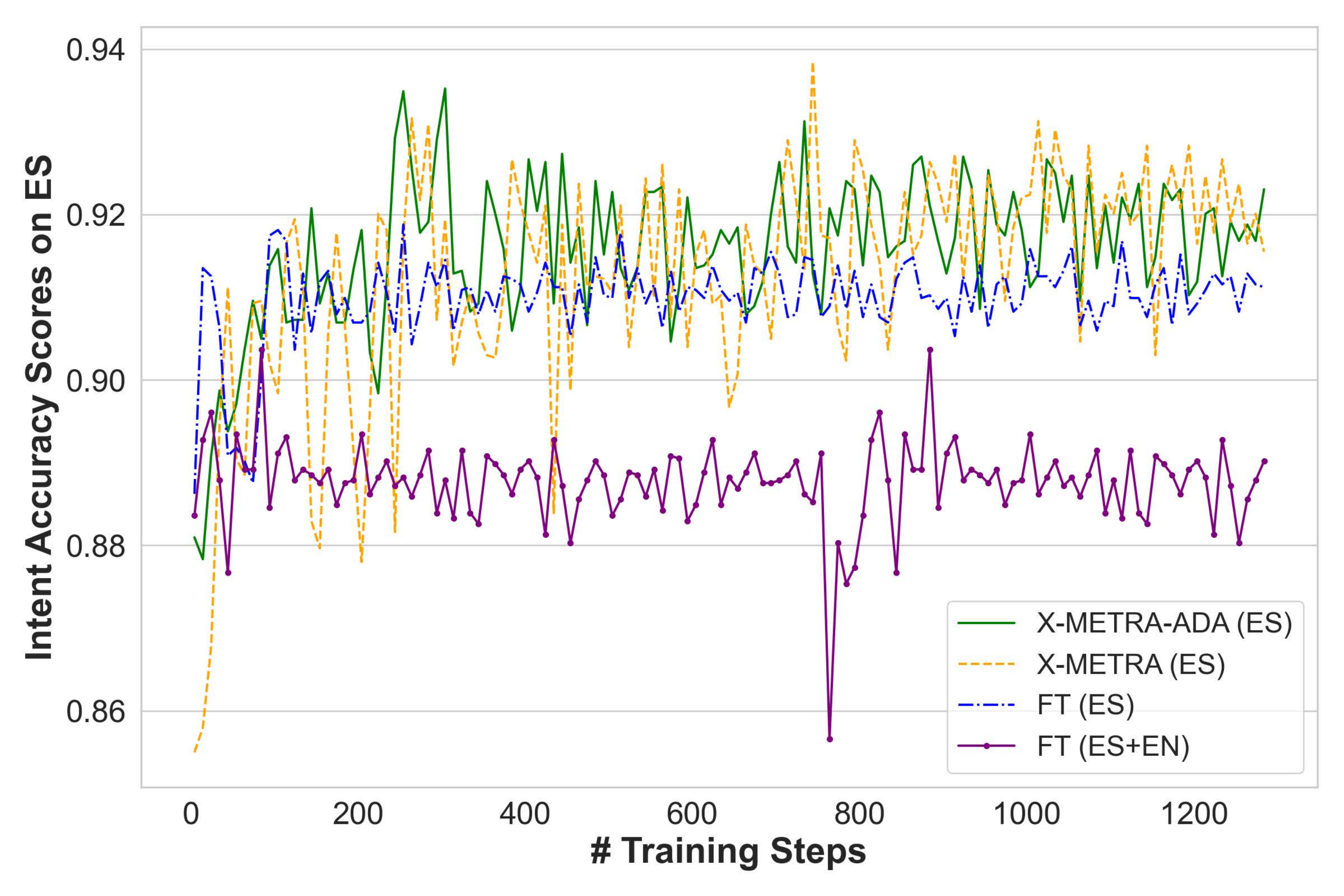}
 \caption{Intent Accuracy on Spanish}
 \label{fig:intent-es}
\end{subfigure}
\begin{subfigure}[b]{0.48\textwidth}
 \centering
 \includegraphics[width=1\textwidth]{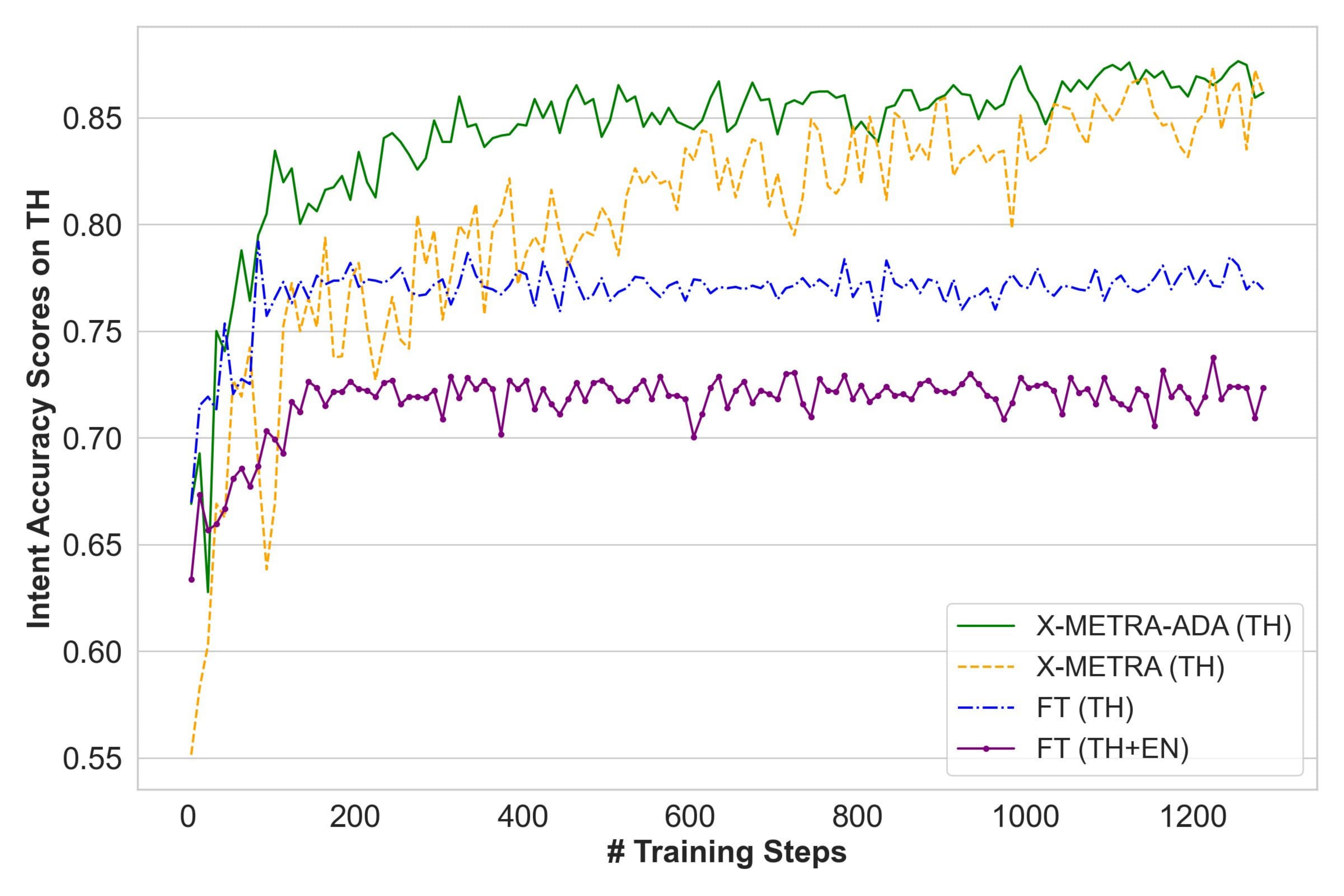}
 \caption{Intent Accuracy on Thai}
 \label{fig:intent-th}
\end{subfigure}
\caption{\label{fig:ablation-study} Ablation of the role of adaptation in X-METRA-ADA compared to X-METRA (X-METRA-ADA with the meta-training stage only). X-METRA-ADA converges faster than X-METRA which in turn is better than FT for both languages. More plots can be found in Appendix \ref{app-ablation}.}
\end{figure*}

Table~\ref{facebook-results-meta} shows the results for cross-lingual transfer learning on MTOD comparing different baselines.\footnote{More results on our in-house NLU dataset can be found in Appendix \ref{app-res-inhouse}.} In general, PRE model performs worse than other baselines. It performs less than the simplest baseline, MCoVe, when transferring to Thai with a decrease of $25.3\%$ and $23.1\%$ and an average cross-lingual relative loss of $4.5\%$ and $2.1\%$ for intent classification and slot filling respectively. This suggests that zero-shot fine-tuning M-BERT on English only is over-fitting on English and its similar languages. Using \textit{MLT$_{A}$} which adds more dialogue-specific mixed training helps reduce that gap for Thai on intent accuracy mainly, but not with the same degree on slot filling. 

The results confirm the positive effects of cross-lingual fine-tuning; although PRE is not a very effective cross-lingual learner, fine-tuning with in-language data on top of PRE (i.e. FT) adds value over the MONO baseline. Adding English data to fine-tuning (FT w/EN) is slightly harmful. However, the meta-learning approach appears to make the most effective use of this data in almost all cases (Spanish slot filling is an exception). We perform a pairwise two-sample t-test (assuming unequal variance) and find the results of X-METRA-ADA compared to FT on intent classification to be statistically significant with p-values of $1.5\%$ and $2.4\%$ for Spanish and Thai respectively, rejecting the null hypothesis with $95\%$ confidence.

X-METRA-ADA outperforms all previous external baselines and fine-tuning models for both Spanish and Thai (except for slot filling on Spanish). We achieve the best overall performance with an average cross-lingual cross-task increase of $3.2\%$ over the FT baseline, $6.9\%$ over FT w/EN, and $12.6\%$ over MONO. Among all models, MONO has the least stability as suggested by higher average standard deviation. There is a tendency for X-METRA-ADA to work better for languages like Thai compared to Spanish as Thai is a truly low-resource language. This suggests that pre-training on English only learns an unsuitable initialization, impeding its generalization to other languages. As expected, fine-tuning on small amounts of the $\dev$ data does not help the model generalize to new languages.
MONO baselines exhibit less stability than X-METRA-ADA. On the other hand, X-METRA-ADA learns a more stable and successful adaptation to that language even on top of a model pre-trained on English with less over-fitting.
 
Table~\ref{qa-meta-short} shows a comparison of methods for TyDiQA-GoldP across seven language, evaluating using F1.\footnote{Full results using Exact Match scores too can be found in Appendix~\ref{app-full-qa}.} The benefits of fine-tuning and improvements from X-METRA-ADA observed in Table~\ref{facebook-results-meta} are confirmed. We also compare X-METRA-ADA to X-METRA, which is equivalent to X-METRA-ADA without the meta-adaptation phase. On average, X-METRA increases by $10.8\%$ and $1.5\%$ over the best external and fine-tuning baseline respectively, whereas MONO results lag behind. X-METRA-ADA outperforms X-METRA on average and is especially helpful on languages like Bengali and Telugu. We compare X-METRA and X-METRA-ADA in more depth in Section~\ref{ana-results}. Meta-learning significantly and consistently outperforms fine-tuning. 

In Appendix \ref{app-full-qa}, we report zero-shot results for QA and notice improvements using X-METRA-ADA over FT for some languages. However, we cannot claim that there is a direct correlation between the degree to which the language is low-resource and the gain in performance of X-METRA-ADA over fine-tuning. Other factors like similarities of grammatical and morphological structure, and shared vocabulary in addition to consistency of annotation may play a role in the observed cross-lingual benefits. Studying such correlations is beyond the scope of this paper.

\subsection{More Analysis}
\label{ana-results}

\paragraph{Meta-Adaptation Role}
The learning curves in Figure~\ref{fig:ablation-study} compare X-METRA-ADA, X-METRA (i.e. meta-training but no meta-adaptation), and fine-tuning, both with English and with target language data only, for both Spanish and Thai intent detection in MTOD. In general, including English data in with in-language fine-tuning data lags behind language-specific training for all models, languages, and sub-tasks. With the exception of slot filling on Spanish, there is a clear gap between naive fine-tuning and meta-learning, with a gain in the favor of X-METRA-ADA especially for Thai. Naive fine-tuning, X-METRA, and X-METRA-ADA all start from the same checkpoint fine-tuned on English. All model variants are sampled from the same data. For Spanish, continuing to use English in naive fine-tuning to Spanish reaches better performance than both variants of meta-learning for Slot filling on Spanish (see Appendix \ref{app-ablation}). This could be due to the typological similarity of Spanish and English, which makes optimization fairly easy for naive fine-tuning compared to Thai, which is both typologically distant and low-resource.
\vspace{-0.2cm}
\paragraph{K-Shot Analysis}
\begin{figure}[h!]
 \includegraphics[height=0.3\textwidth, width=0.5\textwidth]{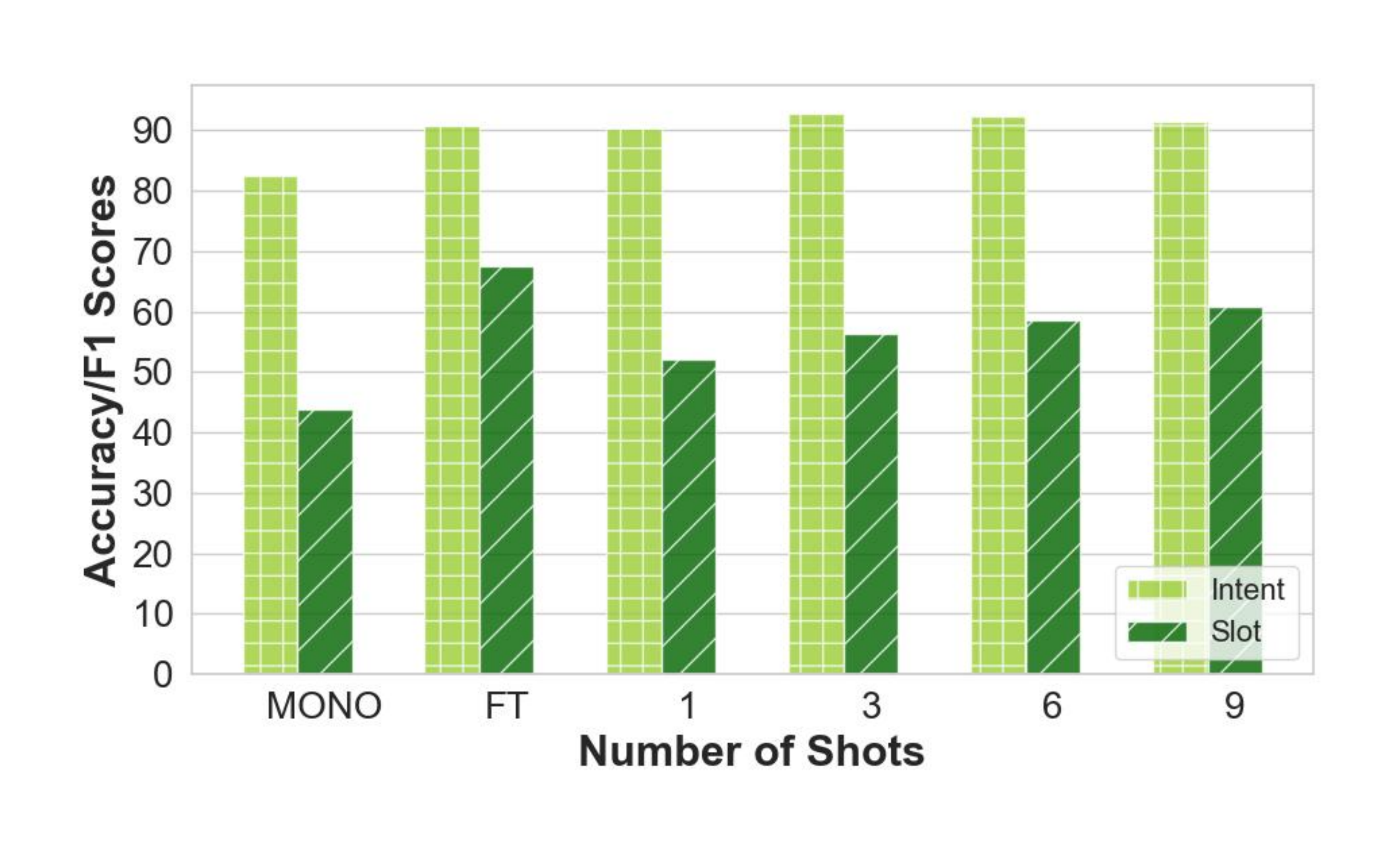}
 \label{fig:k-shot-analysis}
 \vspace{-0.8cm}
 \caption{\label{fig:scala-ana}MTOD intent classification and slot filling on Spanish with different shots. The number of shots is the same for both support and query sets  (i.e. $k=q$).}
\end{figure}
\begin{figure}[h!]
 \includegraphics[height=0.3\textwidth, width=0.5\textwidth]{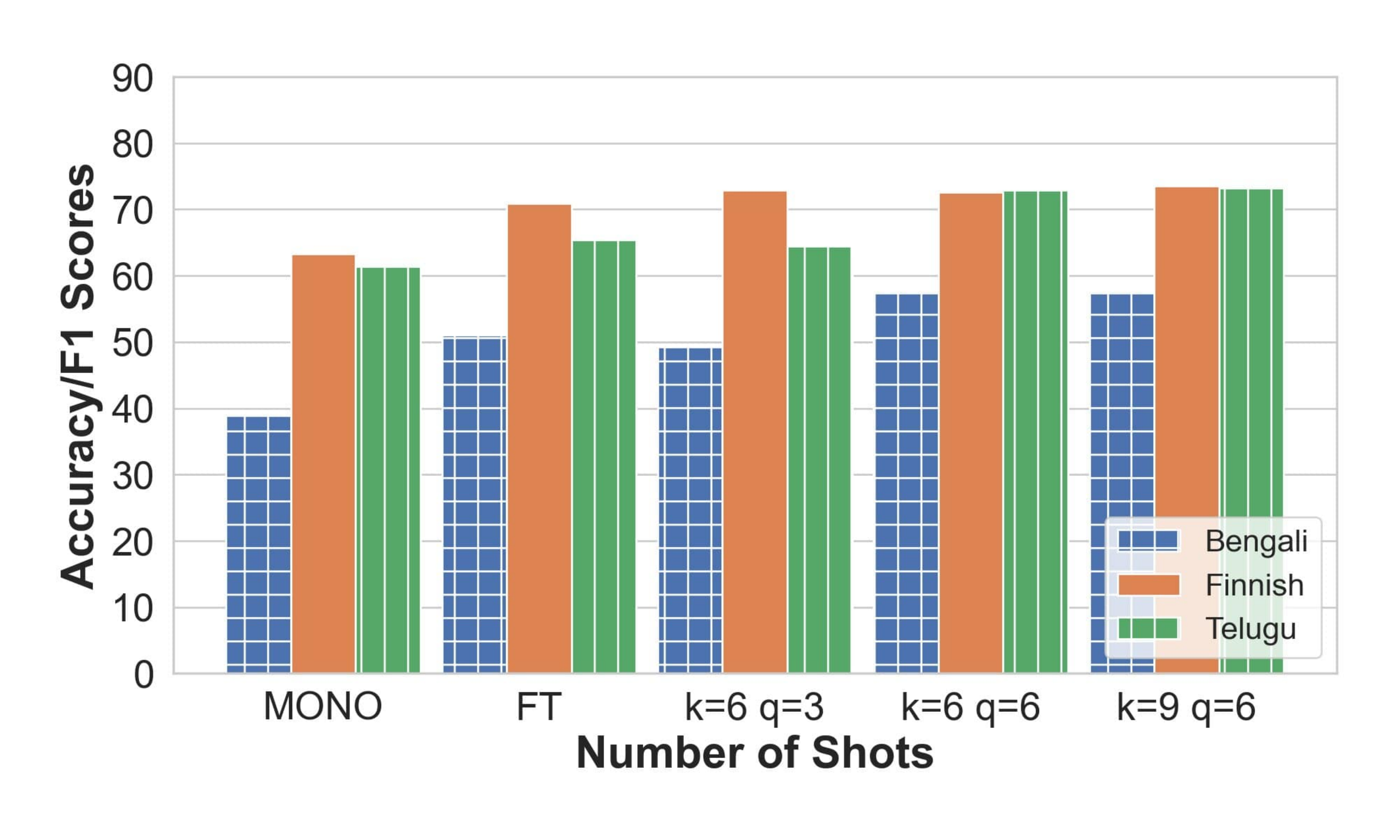}
 \label{fig:k-shot-analysis}
 \vspace{-0.8cm}
 \caption{\label{fig:k-shot-qa} TyDiQA-GoldP F1 score analysis of different shots for both the support and query.}
\end{figure}
We perform a k-shot analysis by treating the number of instances seen per class (i.e. `shots') as a hyper-parameter to determine at which level few-shot meta-learning starts to outperform the fine-tuning and monolingual baselines. As shown in Figure \ref{fig:scala-ana}, it seems that while even one shot for X-METRA-ADA is better than fine-tuning on intent classification, $k=q=9$ shot and $k=q=6$ shot are at the same level of stability with very slightly better results for 6 shot showing that more shots beyond this level will not improve the performance. While 1 shot performance is slightly below our monolingual baseline, it starts approaching the same level of performance as 3 shot upon convergence.

Figure \ref{fig:k-shot-qa} shows an analysis over both $k$ and $q$ shots for TyDiQA-GoldP.  In general, increasing $q$ helps more than increasing $k$. The gap is bigger between $k=6$ $q=3$ and $k=6$ $q=6$ especially for languages like Bengali and Telugu. We can also see that $k=6$ $q=3$ is at the same level of performance to FT for those languages.

\vspace{-0.2cm}

\paragraph{Downsampling Analysis}
\begin{figure}[h!]
 \includegraphics[width=0.5\textwidth]{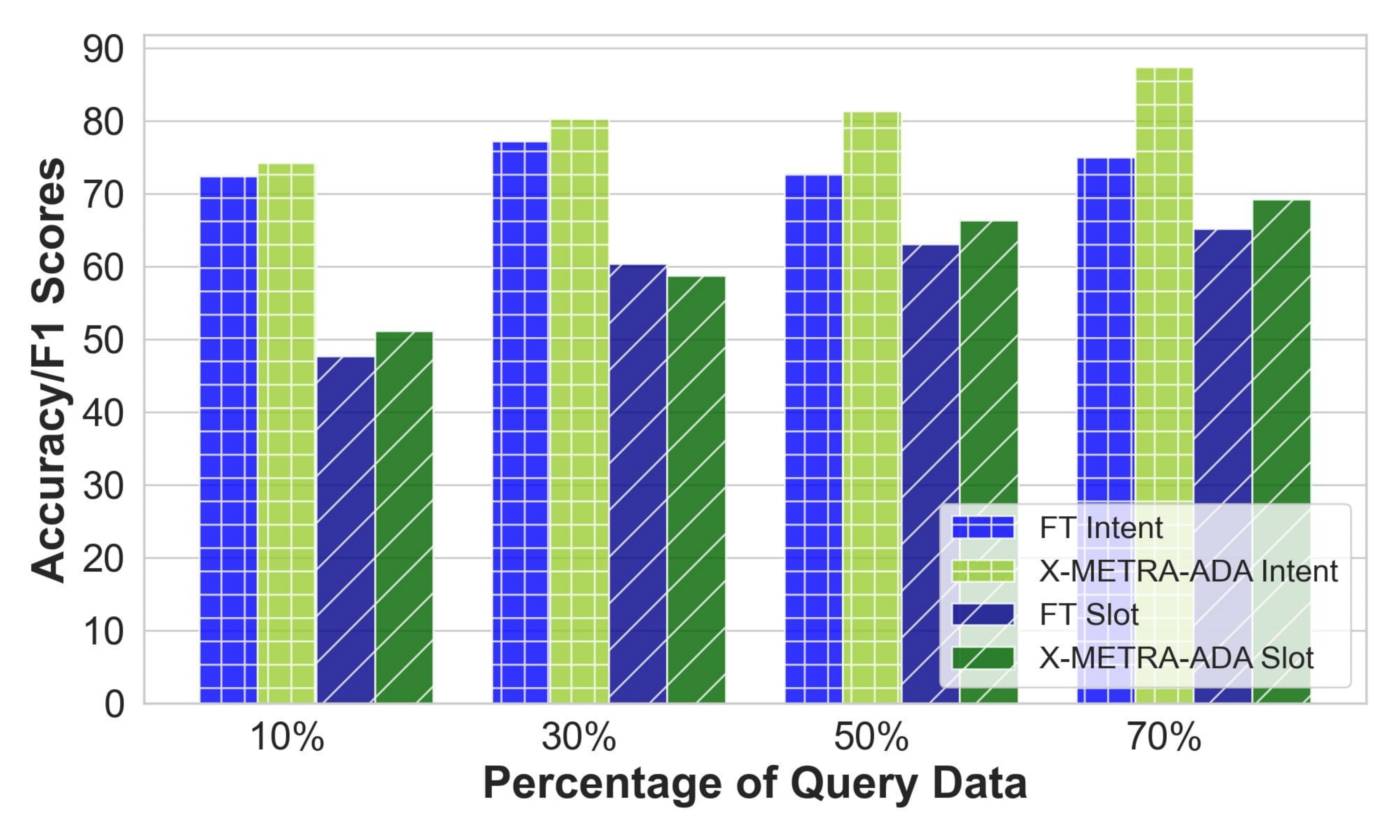}
 \caption{\label{fig:down-th}Downsampling analysis for Thai MTOD with different percentages of query data.}
\end{figure}

We perform a downsampling analysis, where we gradually decrease the proportion of the overall set from which the target language is sampled used for few-shot learning in X-METRA-ADA and FT. Figure  \ref{fig:down-th} shows a comparison between intent accuracies and slot F1 scores between the main models X-METRA-ADA and FT on Thai. We notice that as the percentage of query data increases, the gap between X-METRA-ADA and FT increases slightly, whereas the gain effect on slots is steadier. This suggests that X-METRA-ADA is at the same level of effectiveness even for lower percentages.

\section{Related Work}
\vspace{-0.2cm}

\noindent
\textbf{Cross-lingual transfer learning~~}
Recent efforts apply cross-lingual transfer to downstream applications such as information retrieval ~\cite{jiang-xinfretrieval-lrec-20}; information extraction (\citealp{mhamdi-xeventdet-conll19}, \citealp{bari-zeroner-aaai20}), and chatbot applications (\citealp{lin-xpersona-arxiv20}, ~\citealp{mhamdi-xchurndet-conll18}). \citet{spokenNLU-upadhyay-ieee18} and~\citet{xlingualnlu-schuster-naacl19} propose the first real attempts at cross-lingual task-oriented dialog using transfer learning. Although they show that cross-lingual joint training outperforms monolingual training, their zero-shot model lags behind machine translation for other languages. 

To circumvent imperfect alignments in the cross-lingual representations,~\citet{latentNLU-liu-emnlp19} propose a latent variable model combined with cross-lingual refinement with a small bilingual dictionary related to the dialogue domain.~\citet{mixedNLU-zihan-aaai20} enhance Transformer-based embeddings with mixed language training to learn inter-lingual semantics across languages. However, although these approaches show promising zero-shot performance for Spanish, their learned refined alignments are not good enough to surpass machine translation baselines on Thai. 

More recently,~\citet{xtreme-hu-arxiv20} and ~\citet{xglue-liang-arxiv20} introduce XTREME and XGLUE benchmarks for the large-scale evaluation of cross-lingual capabilities of pre-trained models across a diverse set of understanding and generation tasks. In addition to M-BERT, they analyze models like XLM~\cite{xlm-lample-neurips19} and Unicoder~\cite{unicoder-huang-emnlp19}. Although the latter two models slightly outperform M-BERT, they need a large amount of parallel data to be pre-trained. It is also not clear the extent to which massive cross-lingual supervision helps to bridge the gap to linguistically distant languages.

\noindent
\textbf{Meta-learning for NLP~~}
Previous work in meta-learning for NLP is focused on the application of first-order MAML~\cite{maml-finn-icml17}. Earlier work by~\citet{metamt-gu-emnlp18} extends MAML to improve low-resource languages for neural machine translation. ~\citet{dou-etal-2019-investigating} apply MAML to NLU tasks in the GLUE benchmark. They show that meta-learning is a better alternative to multi-task learning, but they only validate their approach on English.~\citet{xlingualnermaml-wu-aaai20} also use MAML for cross-lingual NER with a slight enhancement to the loss function. More recently,~\citet{xlingualnlumaml-nooralahzadeh-arxiv20} also directly leverage MAML on top of M-BERT and XLM-R for zero-shot and few-shot XNLI and MLQA datasets. Although their attempt shows that cross-lingual transfer using MAML outperforms other baselines, the degree of typological commonalities among languages plays a significant role in that effect. In addition to that, their approach is an oversimplification of the n-way k-shot setup, with a one-fit-all sampling of data points for support and query and additional supervised fine-tuning. 
\vspace{-0.2cm}
\section{Conclusion}
\vspace{-0.1cm}
In this paper, we adapt a meta-learning approach for cross-lingual transfer learning in Natural Language Understanding tasks. Our experiments cover two challenging cross-lingual benchmarks: task-oriented dialog and natural questions including an extensive set of low-resource and typologically diverse languages. X-METRA-ADA reaches better convergence stability on top of fine-tuning, reaching a new state of the art for most languages. 

\section{Acknowledgments}
This work was started while the first author was a research intern at Adobe Research (Summer 2020). This material is partially based upon work supported in part by the Office of the Director of National Intelligence (ODNI), Intelligence Advanced Research Projects Activity (IARPA), via Contract No. 2019-19051600007. The views and conclusions contained herein are those of the authors and should not be interpreted as necessarily representing the official policies, either expressed or implied, of ODNI, IARPA, or the U.S. Government. The U.S. Government is authorized to reproduce and distribute reprints for governmental purposes notwithstanding any copyright annotation therein. We thank the anonymous reviewers for their detailed comments.


\bibliographystyle{acl_natbib}
\bibliography{anthology}


\appendix

\section{Dataset Statistics}
\label{app-data-stats}
Tables \ref{data-mtod} and \ref{data-tydiqa} show the statistics of MTOD and TyDiQA respectively per language and split. 
\begin{table}[H]
\small
\centering
\begin{tabular}{lllll}
\toprule
\textbf{Lang} & \textbf{ISO} & \textbf{Train} & \textbf{Dev} & \textbf{Test} \\ \toprule
English       & EN & 30,521         & 4,181        & 8,621         \\ 
Spanish       & ES & 3,617          & 1,983        & 3,043         \\
Thai          & TH & 2,156          & 1,235       & 1,692         \\ \bottomrule
\end{tabular}
\caption{Statistics of MTOD dataset~\cite{xlingualnlu-schuster-naacl19} per language and split.}
\label{data-mtod}
\end{table}
\begin{table}[H]
\small
\centering
\begin{tabular}{lllll}
\toprule
\textbf{Lang} & \textbf{ISO} & \textbf{Train} & \textbf{Dev} & \textbf{Test} \\ \toprule
English       & EN &   3,326       &  370       & 440        \\ 
Arabic       & AR &   13,324        & 1,481         & 921         \\
Bengali         & BN & 2,151      &  239     & 113       \\ 
Finnish         &  FI & 6,169       & 686      &  782      \\ 
Indonesian         & ID &  5,131        &  571     &  565       \\ 
Russian        & RU &  5,841       &  649     &  812      \\ 
Swahili        & SW &   2,479      &  276     &   499     \\ 
Telugu        & TE &   5,006      &  557     &   669     \\ \bottomrule
\end{tabular}
\caption{Statistics of TyDiQA-GoldP dataset per language and split. Korean is excluded due to some encoding issues.}
\label{data-tydiqa}
\end{table}
\section{Hyperparameters}
\label{app-hyperparam}
For MTOD, we fine-tune PRE on English training data. We use a batch size of $32$, a dropout rate of $0.3$, AdamW with a learning rate of $\num{4e-5}$, and $\epsilon$ of $\num{1e-8}$. We train for around $2000$ steps. Beyond that point more training does not reveal necessary, so we perform early stopping at that point. For MONO, using a smaller learning rate of $\num{4e-5}$ helped achieve a good convergence for that model. For all FT experiments, we use the same learning rate of $\num{1e-3}$, which gave a better convergence.

For QA, we use a batch size of $4$, doc stride of $128$, a fixed maximum sequence length of $384$, and a maximum length of questions of $30$ words. We use AdamW optimizer throughout all experiments, which uses weight decay of $\num{1e-3}$, learning rate of $\num{3e-5}$, and a scheduler of $4$ warm-up steps.\footnote{Those hyperparameters are chosen based on \citet{xtreme-hu-arxiv20}.} We fine-tune PRE for $2$ epochs and observe no more gains in performance. For all MONO and FT experiments, we use the same learning rate of $\num{3e-5}$. This is the same optimizer and learning rate used for the outer loops in meta-learning as well.

For X-METRA-ADA and X-METRA, we sample 2500 tasks in total for both MTOD and QA. For each task, we randomly sample $k=q=6$ examples from each intent class to form the support and query sets respectively (we consider all classes not only the intersection across languages). For QA, we use only one support example per query class and 6 query examples as classes. For the inner loop, we use learn2learn pre-built optimizer.  For the outer loop, we use a standard Adam optimizer. In splitting the few-shot set, we use $75\%$ for the meta-training and $25\%$ for the meta-adaptation for MTOD. For QA, we use $60\%$ of the query for meta-train and the remaining for meta-adaptation.

\section{Results on in-House Intent Classification Dataset}
\label{app-res-inhouse}
We perform an extensive evaluation including other languages for intent classification. We use an in-house dataset covering 6 target languages in addition to English. Statistics of train/dev/test splits are shown in Table ~\ref{inhouse-data}. Table \ref{in-house-results-meta} shows a better performance in the favor of X-METRA with an average cross-lingual gain of $13.5\%$ in accuracy over PRE. We notice that few-shot learning on the language of interest leads to the best performance, as indicated by higher numbers on the diagonal in the confusion matrix. Evaluation on more languages shows some complicity trends between languages from the same family. In addition to that, we notice that languages like Japanese and Korean help each other where few-shot on one helps zero-shot on the other by a margin of 15.6 and 5.6 on Korean and Japanese respectively. 

\begin{table}[h]
\centering
\small
\begin{tabular}{llll}
\toprule
\textbf{Lang} & \textbf{Train} & \textbf{Dev} & \textbf{Test} \\ \toprule
English       & 5,438          & 1,814        & 1,814         \\
German        & 1,570          & 526          & 526           \\
French        & 1,082          & 362          & 362           \\
Italian       & 1,082          & 362          & 362           \\
Portuguese    & 1,150          & 386          & 386           \\
Japanese      & 1,070          & 358          & 358           \\
Korean        & 938            & 314          & 314           \\ \bottomrule
\end{tabular}
\caption{Statistics of In-House multilingual intent classification Dataset.}
\label{inhouse-data}
\end{table}

\begin{table}[]
\centering
\scalebox{0.76}{
\begin{tabular}{l|l|l|l|l|l|l|l}  \toprule
\multirow{2}{*}{\textbf{Type}}          & \multirow{2}{*}{\textbf{Model}}   &\multicolumn{6}{l} {\textbf{Test on}}
\\ \cline{3-8} 
                                                      &       &  DE & FR & IT & PT & JA & KR \\ \toprule
\multicolumn{1}{l|}{PRE} & EN & 19.1       & 30.0       & 30.1       & 26.1       & 14.6       & 5.1        \\ \hline
\multicolumn{1}{l|}{\multirow{6}{*}{X-METRA}}     & DE     & \textbf{34.3}       & 33.3       & 30.0       & 30.2       & 13.5       & 8.9        \\ \cline{2-8} 
\multicolumn{1}{l|}{}                          & FR     & 19.2       & \textbf{34.1}       & 29.9       & 29.1       & 5.8        & 9.0        \\ \cline{2-8} 
\multicolumn{1}{l|}{}                          & IT     & 18.3       & 32.2       & \textbf{44.4}       & 30.2       & 6.7        & 10.2       \\ \cline{2-8} 
\multicolumn{1}{l|}{}                          & PT     & 19.1       & 27.7       & 30.1       & \textbf{31.4}       & 5.8        & 9.0        \\ \cline{2-8} 
\multicolumn{1}{l|}{}                          & JA     & 24.1       & 25.7       & 33.2       & 26.1       & \textbf{30.9}       & 20.7       \\ \cline{2-8} 
\multicolumn{1}{l|}{}                          & KR     & 24.4       & 25.6       & 34.4       & 25.0       & 20.2       & \textbf{30.7}       \\ 
\bottomrule
\end{tabular}
}
\caption{X-METRA results on an In-House multilingual intent data. Bold results highlight best results for each test language.\label{in-house-results-meta}}
\end{table}

\section{Full results for QA}
\label{app-full-qa}

Tables~\ref{qa-meta-f1-full} and \ref{qa-meta-em-full} show the full results for F1 and Exact Match (EM) metrics for QA respectively.
\begin{table*}[h!] 
\small
\centering

\begin{tabular}{l|lllllll} \toprule   \multirow{2}{*}{\textbf{Model}}  &\multicolumn{7}{c} {\textbf{Test on}}
\\
& AR & BN & FI & ID  &  RU & SW & TE  \\ \toprule
\multicolumn{8}{l}{\textbf{MONO}}    \\ \midrule  
\hspace{0.2cm} AR  &  74.0 $ \pm 1.1 $ &
30.1 $ \pm 2.4 $ &
50.0 $ \pm 0.8 $ &
59.5 $ \pm 1.3 $ &
48.4 $ \pm 0.8 $ &
50.8 $ \pm 1.7 $ &
24.1 $ \pm 2.7 $ 
 \\ 
\hspace{0.2cm} BN                                                                                           & 32.2 $ \pm 2.6 $ &
38.9 $ \pm 0.8 $ &
33.9 $ \pm 1.4 $ &
36.3 $ \pm 1.5 $ &
31.8 $ \pm 1.4 $ &
37.2 $ \pm 1.8 $ &
34.7 $ \pm 4.2 $ 
      \\ 
\hspace{0.2cm}  FI                                                                                           & 54.2 $ \pm 2.5 $ &
30.7 $ \pm 1.3 $ &
63.3 $ \pm 1.5 $ &
52.5 $ \pm 1.7 $ &
43.0 $ \pm 2.1 $ &
48.6 $ \pm 1.7 $ &
28.7 $ \pm 2.8 $ 
    \\
\hspace{0.2cm} ID                                                                                         &   58.0 $ \pm 1.8 $ &
31.8 $ \pm 0.5 $ &
48.2 $ \pm 2.0 $ &
67.1 $ \pm 1.9 $ &
45.1 $ \pm 1.8 $ &
50.3 $ \pm 1.8 $ &
29.4 $ \pm 2.7 $ 
  \\ 
\hspace{0.2cm} RU                                                                                          &  50.9 $ \pm 2.3 $ &
34.5 $ \pm 2.1 $ &
45.2 $ \pm 4.2 $ &
52.0 $ \pm 4.0 $ &
54.4 $ \pm 1.3 $ &
47.1 $ \pm 2.1 $ &
30.7 $ \pm 2.5 $ 
  \\ 
\hspace{0.2cm} SW                                                                                            & 35.8 $ \pm 1.5 $ &
27.6 $ \pm 1.5 $ &
33.6 $ \pm 2.1 $ &
37.4 $ \pm 1.9 $ &
25.7 $ \pm 1.7 $ &
60.3 $ \pm 1.2 $ &
13.2 $ \pm 2.3 $ 
   \\ 
\hspace{0.2cm} TE                                                                                            &34.0 $ \pm 0.9 $ &
38.0 $ \pm 2.2 $ &
39.5 $ \pm 0.6 $ &
35.3 $ \pm 1.1 $ &
35.9 $ \pm 1.1 $ &
43.5 $ \pm 1.0 $ &
61.4 $ \pm 1.0 $ \\ \hline
\multicolumn{8}{l}{\textbf{FT}}    \\ \midrule 
\hspace{0.2cm} AR                                 &   \underline{77.0} $ \pm 0.3 $ &
36.8 $ \pm 2.9 $ &
58.8 $ \pm 0.6 $ &
67.0 $ \pm 2.7 $ &
60.9 $ \pm 0.8 $ &
52.4 $ \pm 3.6 $ &
32.0 $ \pm 1.0 $ 
\\                                                  
\hspace{0.2cm} BN                                                                                           &  60.7 $ \pm 0.4 $ &
51.0 $ \pm 2.7 $ &
59.2 $ \pm 0.6 $ &
67.1 $ \pm 1.6 $ &
59.2 $ \pm 0.3 $ &
56.2 $ \pm 0.8 $ &
43.7 $ \pm 0.9 $  \\ 
\hspace{0.2cm} FI                                                                                           &   60.3 $ \pm 1.9 $ &
36.7 $ \pm 1.3 $ &
70.9 $ \pm 0.4 $ &
65.7 $ \pm 1.4 $ &
62.1 $ \pm 0.5 $ &
50.9 $ \pm 1.3 $ &
36.4 $ \pm 3.6 $ 
       \\ 
\hspace{0.2cm} ID                                                                                   &  65.7 $ \pm 1.4 $ &
37.0 $ \pm 1.1 $ &
60.8 $ \pm 0.2 $ &
77.0 $ \pm 0.4 $ &
61.1 $ \pm 0.5 $ &
56.8 $ \pm 1.0 $ &
36.7 $ \pm 0.4 $ 
  \\ 
\hspace{0.2cm} RU                                                                                            &   60.9 $ \pm 2.5 $ &
37.2 $ \pm 2.0 $ &
59.0 $ \pm 2.1 $ &
66.8 $ \pm 1.3 $ &
64.8 $ \pm 0.4 $ &
55.2 $ \pm 1.8 $ &
36.8 $ \pm 1.3 $ 
  \\ 
\hspace{0.2cm} SW                                                                                            & 57.4 $ \pm 0.5 $ &
35.2 $ \pm 1.5 $ &
56.2 $ \pm 1.0 $ &
65.4 $ \pm 1.8 $ &
58.8 $ \pm 0.8 $ &
70.2 $ \pm 1.7 $ &
33.1 $ \pm 2.8 $ 
    \\ 
\hspace{0.2cm} TE                                                                                            &    54.0 $ \pm 3.2 $ &
39.1 $ \pm 2.1 $ &
54.8 $ \pm 2.3 $ &
63.5 $ \pm 2.6 $ &
58.1 $ \pm 0.9 $ &
56.9 $ \pm 1.8 $ &
65.4 $ \pm 0.6 $ 
 \\ \midrule 
 
 \multicolumn{8}{l}{\textbf{X-METRA}}    \\ \midrule   \hspace{0.2cm}  AR                                                                                        & \textbf{78.4} $ \pm 0.6 $ &
33.0 $ \pm 0.8 $ &
58.2 $ \pm 0.2 $ &
66.4 $ \pm 1.4 $ &
59.9 $ \pm 0.1 $ &
53.2 $ \pm 3.8 $ &
31.4 $ \pm 3.0 $ 
 \\  
\hspace{0.2cm} BN                                                                                         & 56.9 $ \pm 3.2 $ &
\underline{53.2} $ \pm 0.5 $ &
56.7 $ \pm 1.4 $ &
67.4 $ \pm 1.2 $ &
56.7 $ \pm 1.3 $ &
56.0 $ \pm 0.9 $ &
41.7 $ \pm 0.6 $ 
 \\  
\hspace{0.2cm} FI                                                                                           &    58.9 $ \pm 0.6 $ &
33.6 $ \pm 1.1 $ &
\underline{72.8} $ \pm 0.3 $ &
61.9 $ \pm 2.0 $ &
60.7 $ \pm 0.9 $ &
46.5 $ \pm 1.2 $ &
36.6 $ \pm 1.7 $ 
  \\ 
\hspace{0.2cm} ID                                                                                          &     65.8 $ \pm 0.3 $ &
35.0 $ \pm 2.2 $ &
60.5 $ \pm 0.9 $ &
\textbf{77.7} $ \pm 0.2 $ &
60.4 $ \pm 1.3 $ &
57.4 $ \pm 1.1 $ &
35.3 $ \pm 0.3 $     
\\ 
\hspace{0.2cm} RU                                                                                          &  60.3 $ \pm 1.6 $ &
37.2 $ \pm 0.7 $ &
59.1 $ \pm 0.3 $ &
66.8 $ \pm 0.8 $ &
\textbf{66.2} $ \pm 0.1 $ &
53.7 $ \pm 0.8 $ &
33.2 $ \pm 3.1 $ 
 \\ 
\hspace{0.2cm} SW                                                                                         &  58.5 $ \pm 0.0 $ &
36.9 $ \pm 1.2 $ &
56.0 $ \pm 0.2 $ &
64.8 $ \pm 0.7 $ &
58.4 $ \pm 0.4 $ &
\textbf{71.9} $ \pm 0.2 $ &
33.7 $ \pm 1.5 $  \\

\hspace{0.2cm} TE                                                                                           & 56.0 $ \pm 3.0 $ &
38.8 $ \pm 0.1 $ &
53.6 $ \pm 1.7 $ &
61.1 $ \pm 1.9 $ &
58.6 $ \pm 0.6 $ &
55.8 $ \pm 0.2 $ &
\underline{66.4} $ \pm 0.5 $ 
 \\ \midrule
 
\multicolumn{8}{l}{\textbf{X-METRA-ADA}}    \\ \midrule    \hspace{0.2cm} AR                                                                                        & 76.6 $ \pm 0.1 $ &
 49.6 $ \pm  1.3$ &
 63.4 $ \pm  0.4 $ &
70.9 $ \pm 0.1 $ &
60.1 $ \pm  1.0$ &
56.8 $ \pm  0.4 $ &
42.4 $ \pm  2.5$ 
 \\ 
\hspace{0.2cm} BN         &                                                                     59.4 $ \pm  0.3$ &
\textbf{57.8} $ \pm 0.6$ &
59.2 $ \pm 0.2$ &
63.1 $ \pm  0.2$ &
56.5 $ \pm  0.2$ &
56.1 $ \pm  0.3$ &
44.1 $ \pm 0.4 $ 
 \\ 
\hspace{0.2cm} FI                                                                                          &  62.8 $ \pm 1.3$ &
50.8 $ \pm 1.3$ &
\textbf{73.0} $ \pm  0.3$ &
65.5 $ \pm  1.2$ &
60.1 $ \pm  0.4$ &
54.9 $ \pm  0.3$ &
42.5 $ \pm  0.5$ 
  \\ 
\hspace{0.2cm} ID  & 
66.7  $ \pm 0.3$ &
49.9 $ \pm 0.5$ &
62.6 $ \pm 0.7$ &
\underline{77.3} $ \pm 0.1$ &
58.3 $ \pm 0.9$ &
58.1 $ \pm 0.6$ &
42.6 $ \pm 0.4$    
\\
\hspace{0.2cm} RU                                                                                          & 62.2 $ \pm 0.7$ &
47.6 $ \pm 1.6$ &
63.1 $ \pm 0.2$ &
63.4 $ \pm 0.9$ &
\textbf{66.9} $ \pm 0.1$ &
56.0 $ \pm 1.1$ &
43.3 $ \pm 1.2$  
 \\   
\hspace{0.2cm} SW                                                                                         &  59.1 $ \pm 0.7$ &
49.1 $ \pm 1.1$ &
58.1 $ \pm 0.2$ &
62.1 $ \pm 1.0$ &
54.6 $ \pm 0.6$ &
\underline{70.3} $ \pm 0.2$ &
43.2 $ \pm 0.7$ 
  \\

\hspace{0.2cm} TE                                                                                           & 58.2 $ \pm 2.8$ &
52.1 $ \pm 1.7$ &
61.5 $ \pm 1.0$ &
62.0 $ \pm 0.5$ &
58.2 $ \pm 0.5$ &
59.7 $ \pm 1.4$ &
\textbf{72.8} $ \pm 0.1$  
 \\ 
\bottomrule
\end{tabular}
\caption{\label{qa-meta-f1-full} Full F1 Results on TyDiQA-GoldP between external, pre-training, monolingual and fine-tuning baselines on one hand and X-METRA and X-METRA-ADA on the other hand.}
\end{table*}

\begin{table*}[h] 
\small
\centering
\begin{tabular}{l|l|lllllll} \toprule
\multirow{2}{*}{\textbf{Model}}  &\multicolumn{7}{c} {\textbf{Test on}} 
\\  & AR & BN & FI & ID  &  RU & SW & TE &  
 \\ \midrule 
 
 \multicolumn{8}{l}{\textbf{MONO}}    \\ \midrule
\hspace{0.2cm} AR   & 
57.5 $ \pm 1.5 $ &
19.7 $ \pm 2.9 $ &
35.1 $ \pm 1.0 $ &
44.2 $ \pm 1.3 $ &
25.2 $ \pm 0.9 $ &
33.8 $ \pm 1.4 $ &
14.9 $ \pm 1.7 $ 

\\                                                 
\hspace{0.2cm} BN   & 
17.1 $ \pm 1.4 $ &
24.5 $ \pm 2.9 $ &
17.5 $ \pm 0.4 $ &
20.8 $ \pm 2.0 $ &
14.4 $ \pm 0.5 $ &
20.5 $ \pm 1.4 $ &
19.9 $ \pm 5.0 $ 

\\ 
\hspace{0.2cm} FI  & 
33.7 $ \pm 4.0 $ &
15.6 $ \pm 1.6 $ &
49.8 $ \pm 1.3 $ &
35.3 $ \pm 2.3 $ &
21.4 $ \pm 1.4 $ &
26.1 $ \pm 9.9 $ &
16.5 $ \pm 3.9 $ 

\\ 
\hspace{0.2cm} ID & 
39.7 $ \pm 1.4 $ &
18.6 $ \pm 1.3 $ &
32.7 $ \pm 1.9 $ &
54.9 $ \pm 0.1 $ &
23.8 $ \pm 0.6 $ &
34.4 $ \pm 1.2 $ &
16.9 $ \pm 4.9 $

\\ 
\hspace{0.2cm} RU  & 
30.8 $ \pm 1.9 $ &
26.3 $ \pm 4.9 $ &
29.7 $ \pm 2.4 $ &
34.9 $ \pm 4.0 $ &
37.9 $ \pm 1.6 $ &
30.7 $ \pm 3.1 $ &
19.9 $ \pm 1.9 $

\\ 
\hspace{0.2cm} SW   & 
16.0 $ \pm 1.3 $ &
16.5 $ \pm 1.5 $ &
15.6 $ \pm 1.0 $ &
21.1 $ \pm 1.3 $ &
10.5 $ \pm 1.3 $ &
48.6 $ \pm 1.2 $ &
5.3 $ \pm 1.7 $ 

\\ 
\hspace{0.2cm} TE  & 
18.8 $ \pm 2.0 $ &
26.3 $ \pm 1.5 $ &
23.8 $ \pm 2.6 $ &
21.6 $ \pm 2.5 $ &
20.4 $ \pm 1.2 $ &
26.7 $ \pm 1.7 $ &
46.3 $ \pm 1.1 $ 

\\ \midrule 
\multicolumn{8}{l}{\textbf{FT}}    \\ \midrule 

\hspace{0.2cm} AR & 
\underline{61.3} $ \pm 1.0 $ &
26.5 $ \pm 4.4 $ &
43.1 $ \pm 1.0 $ &
52.2 $ \pm 2.0 $ &
37.9 $ \pm 2.5 $ &
35.6 $ \pm 3.3 $ &
21.0 $ \pm 3.0 $

\\                                  \hspace{0.2cm}             BN  & 
42.2 $ \pm 0.9 $ & 
38.0 $ \pm 4.4 $ &
44.8 $ \pm 1.2 $ &
51.5 $ \pm 2.2 $ &
36.8 $ \pm 1.6 $ &
37.2 $ \pm 1.7 $ &
27.3 $ \pm 0.2 $ 

\\ 
\hspace{0.2cm} FI & 
43.2 $ \pm 1.8 $ &
23.6 $ \pm 1.1 $ &
56.5 $ \pm 0.6 $ &
50.8 $ \pm 2.1 $ &
40.5 $ \pm 0.8 $ &
33.5 $ \pm 1.2 $ &
20.7 $ \pm 3.3 $

\\ 
\hspace{0.2cm} ID  & 
49.4 $ \pm 1.6 $ &
23.3 $ \pm 2.4 $ &
46.4 $ \pm 0.4 $ &
\underline{63.8} $ \pm 0.5 $ &
40.5 $ \pm 0.1 $ &
38.1 $ \pm 2.1 $ &
24.1 $ \pm 0.5 $

\\ 
\hspace{0.2cm} RU  & 
42.6 $ \pm 2.6 $ &
24.8 $ \pm 3.3 $ &
43.5 $ \pm 2.0 $ &
52.4 $ \pm 2.3 $ &
46.5 $ \pm 0.4 $ &
37.6 $ \pm 1.5 $ &
24.5 $ \pm 1.3 $ 

\\ 
\hspace{0.2cm} SW  & 
38.9 $ \pm 0.6 $ &
23.0 $ \pm 1.4 $ &
40.1 $ \pm 1.4 $ &
50.0 $ \pm 1.7 $ &
38.0 $ \pm 0.8 $ &
59.0 $ \pm 3.1 $ &
23.5 $ \pm 1.4 $

\\
\hspace{0.2cm} TE  & 
36.1 $ \pm 2.2 $ &
30.0 $ \pm 2.3 $ &
40.0 $ \pm 2.5 $ &
49.4 $ \pm 2.1 $ &
38.6 $ \pm 0.9 $ &
39.0 $ \pm 1.7 $ &
49.2 $ \pm 0.5 $
 
\\ \midrule 
\multicolumn{8}{l}{\textbf{X-METRA}}    \\ \midrule 

\hspace{0.2cm} AR  & 
\textbf{63.3} $ \pm 0.8 $ &
21.2 $ \pm 1.9 $ &
42.6 $ \pm 1.0 $ &
51.8 $ \pm 1.2 $ &
34.9 $ \pm 1.1 $ &
36.0 $ \pm 3.5 $ &
20.9 $ \pm 1.7 $ 

\\
\hspace{0.2cm} BN    & 
29.2 $ \pm 16.5 $ &
\underline{39.0} $ \pm 1.9 $ &
41.9 $ \pm 1.6 $ &
51.1 $ \pm 1.7 $ &
34.1 $ \pm 0.4 $ &
37.1 $ \pm 1.4 $ &
25.6 $ \pm 0.2 $ 

\\
\hspace{0.2cm} FI  & 
42.0 $ \pm 1.0 $ &
20.4 $ \pm 0.7 $ &
\textbf{59.1} $ \pm 1.1 $ &
46.0 $ \pm 2.7 $ &
36.8 $ \pm 1.3 $ &
30.9 $ \pm 0.6 $ &
22.5 $ \pm 0.9 $ 

\\
\hspace{0.2cm} ID  & 
54.8 $ \pm 7.9 $ &
20.1 $ \pm 1.5 $ &
46.1 $ \pm 1.2 $ &
\textbf{65.2} $ \pm 0.5 $ &
38.5 $ \pm 1.9 $ &
39.6 $ \pm 0.8 $ &
23.1 $ \pm 1.4 $ 

\\ 
\hspace{0.2cm} RU   & 
42.9 $ \pm 1.3 $ &
26.5 $ \pm 1.2 $ &
43.0 $ \pm 0.6 $ &
53.0 $ \pm 0.1 $ &
\textbf{48.9} $ \pm 0.4 $ &
35.3 $ \pm 1.0 $ &
21.6 $ \pm 2.4 $ 

\\
\hspace{0.2cm} SW   &
39.9 $ \pm 0.4 $ &
26.0 $ \pm 1.1 $ &
40.0 $ \pm 0.7 $ &
50.3 $ \pm 0.4 $ &
38.0 $ \pm 0.9 $ &
\textbf{61.4} $ \pm 0.4 $ &
23.9 $ \pm 0.7 $ 

\\ 
\hspace{0.2cm} TE   & 
38.0 $ \pm 3.9 $ &
28.3 $ \pm 0.0 $ &
37.0 $ \pm 2.3 $ &
47.6 $ \pm 3.4 $ &
36.3 $ \pm 0.5 $ &
36.9 $ \pm 1.2 $ &
\underline{49.7} $ \pm 0.5 $ 
\\  \midrule
 
\multicolumn{8}{l}{\textbf{X-METRA-ADA}}    \\ \midrule   \hspace{0.2cm} AR                                                                                        & 55.0 $ \pm 0.3 $ &
36.0 $ \pm 3.0$ &
43.8 $ \pm 0.5$ &
55.2 $ \pm 0.5$ &
35.4 $ \pm 2.6$ &
40.0 $ \pm 0.2$ &
31.9 $ \pm 2.2$ 
 \\ 
\hspace{0.2cm} BN                                                                                         & 38.4 $ \pm 0.3$ &
\textbf{41.0} $ \pm 0.8$ &
 43.5 $ \pm 0.4$ &
46.7 $ \pm  0.1$ &
32.4 $ \pm 0.4$ &
37.9 $ \pm 0.4$ &
33.8 $ \pm 0.7 $ 
 \\   
\hspace{0.2cm} FI                                                                                          & 40.9 $ \pm 1.1$ &
34.2 $ \pm 1.1$ &
\underline{57.9} $ \pm 1.0$ &
49.0 $ \pm 1.4$ &
35.3 $ \pm 0.3$ &
38.0 $ \pm 0.7$ &
30.0 $ \pm 1.0$ 
  \\   
\hspace{0.2cm} ID                                                                                          & 45.4 $ \pm 0.4$ &
33.9 $ \pm 1.1$ &
47.6 $ \pm 0.4$ &
63.4 $ \pm 0.4$ &
36.3 $ \pm 0.9$ &
43.4 $ \pm 0.8$ &
31.9 $ \pm 0.2$    
\\ 
\hspace{0.2cm} RU                                                                                          & 39.4 $ \pm 0.1$ &
34.8 $ \pm 1.5$ &
45.1 $ \pm 0.5$ &
48.6 $ \pm 0.9$ &
\underline{47.5} $ \pm 0.3$ &
39.3 $ \pm 1.3$ &
33.8 $ \pm 1.3$  
 \\  
\hspace{0.2cm} SW                                              & 36.7 $ \pm 0.6$ &
36.3 $ \pm 1.4$ &
42.5 $ \pm 0.5$ &
45.8 $ \pm 1.4$ &
32.4 $ \pm 0.7$ &
\underline{59.6} $ \pm 0.5$ &
33.8 $ \pm 1.0$                     
  \\

\hspace{0.2cm} TE     & 37.9 $ \pm 1.9$ &
38.1 $ \pm 2.6$ &
44.9 $ \pm 1.4$ &
48.0 $ \pm 0.3$ &
38.8 $ \pm 0.4$ &
43.5 $ \pm 1.6$ &
\textbf{56.4} $ \pm 0.4$                                                                                        
 \\ 
\bottomrule
\end{tabular}
\caption{\label{qa-meta-em-full} Full EM Results on TyDiQA-GoldP between external, pre-training, monolingual and fine-tuning baselines on one hand, X-METRA and X-METRA-ADA on the other hand.}
\end{table*}

\section{More Ablation}
\label{app-ablation}

\begin{figure*}
\centering
\begin{subfigure}[b]{0.48\textwidth}
\centering
\includegraphics[width=\textwidth]{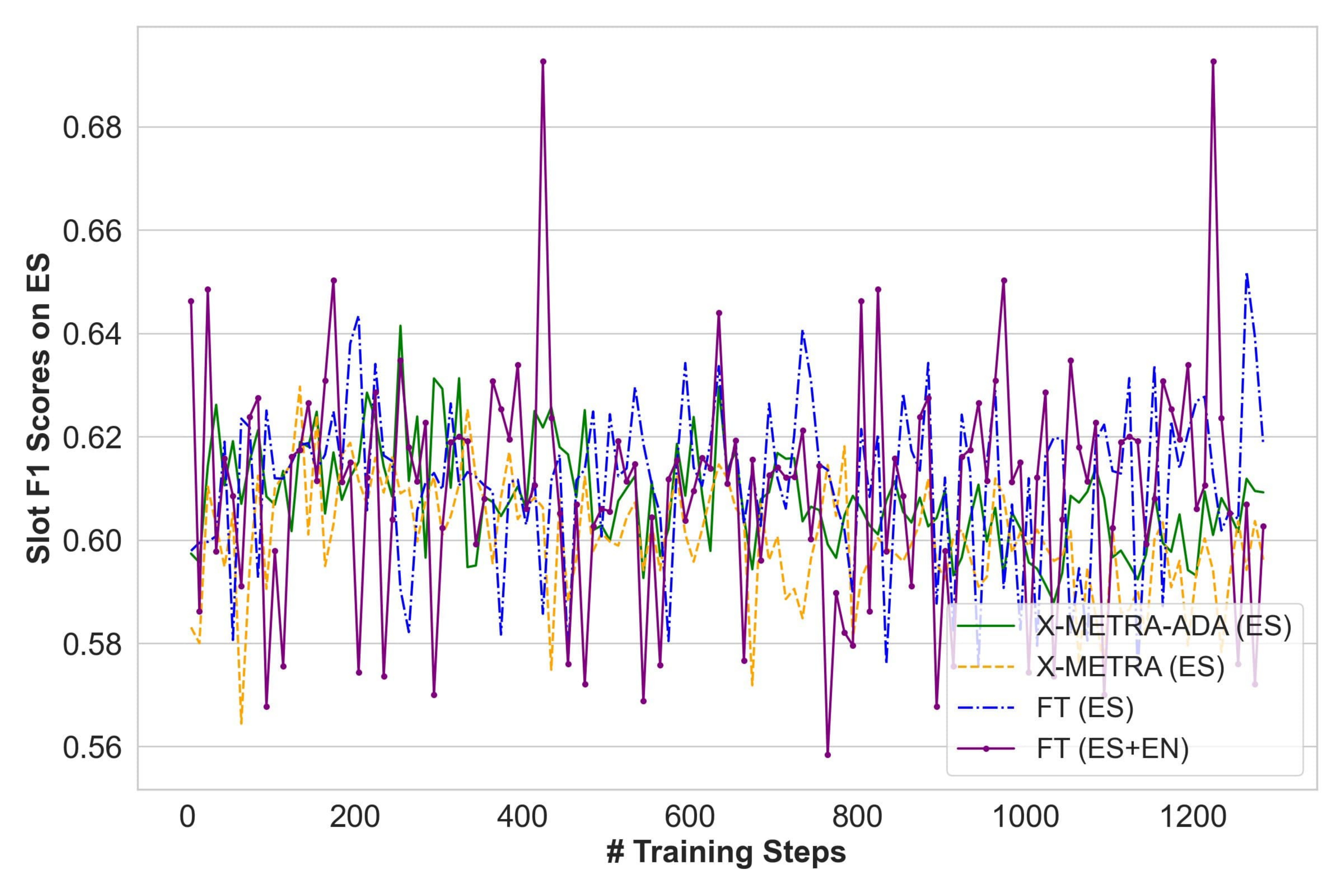}
\caption{Slot F1 on Spanish}
\label{fig:slot-es-abla}
\end{subfigure}
\begin{subfigure}[b]{0.48\textwidth}
 \centering
 \includegraphics[width=1\textwidth]{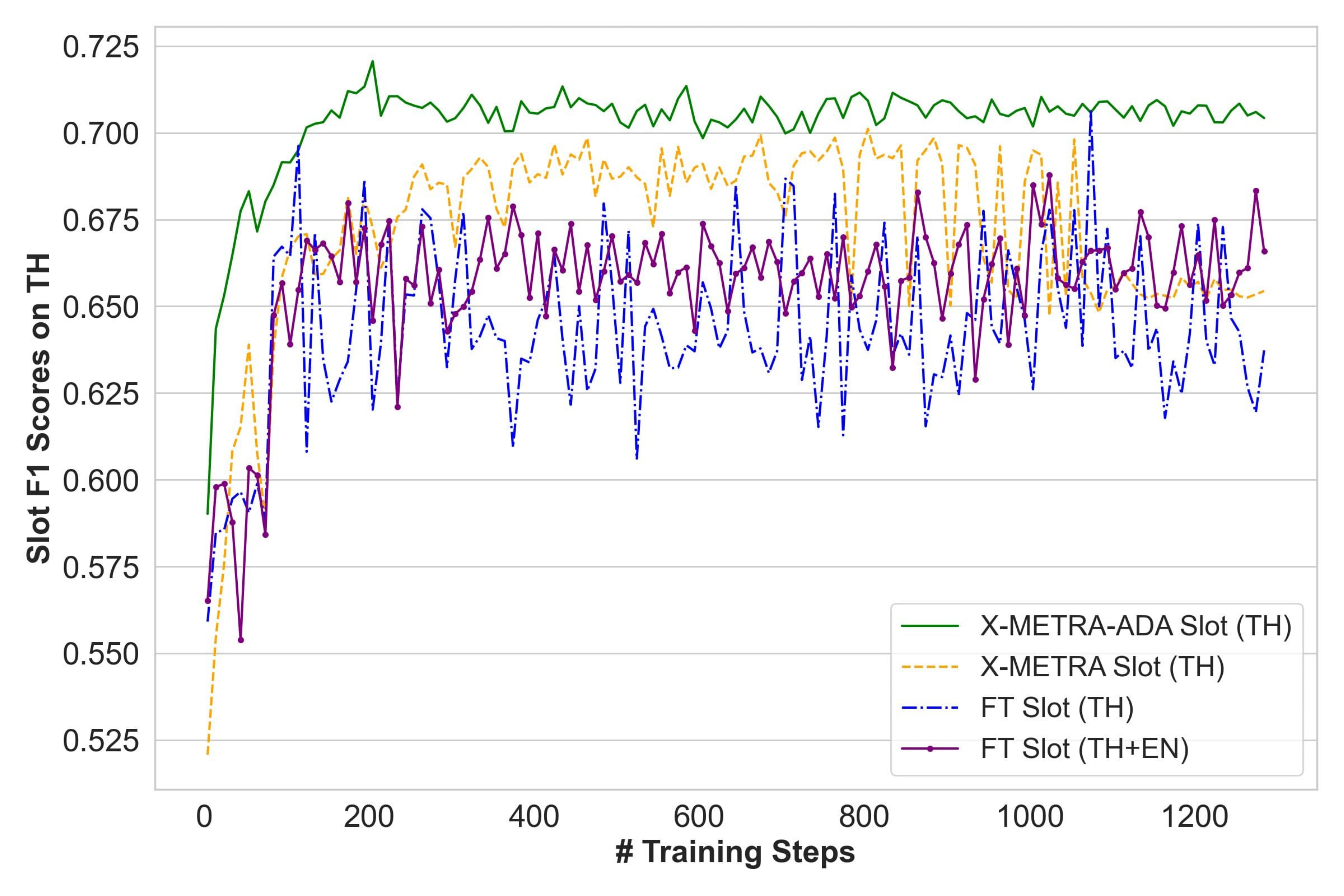}
 \caption{Slot F1 on Thai}
 \label{fig:slot-th-abla}
 \end{subfigure}
 
 \caption{\label{fig:ablation-study-slot} Ablation Study on the role of the adaptation in X-METRA-ADA compared to X-METRA (MAML with only the meta-training stage) for different languages, language-specific vs joint training. All models are compared to their fine-tuning counterparts. }
\end{figure*}

Figure \ref{fig:ablation-study-slot} compares between the learning curves for language-specific and joint training with respect to slot filling for both Spanish and Thai. 

\begin{figure*}

\end{figure*}
\section{More Analysis}
\label{app-more-analysis}
\paragraph{More Downsampling Analysis}

\begin{figure}[h]{}
 \includegraphics[width=0.5\textwidth]{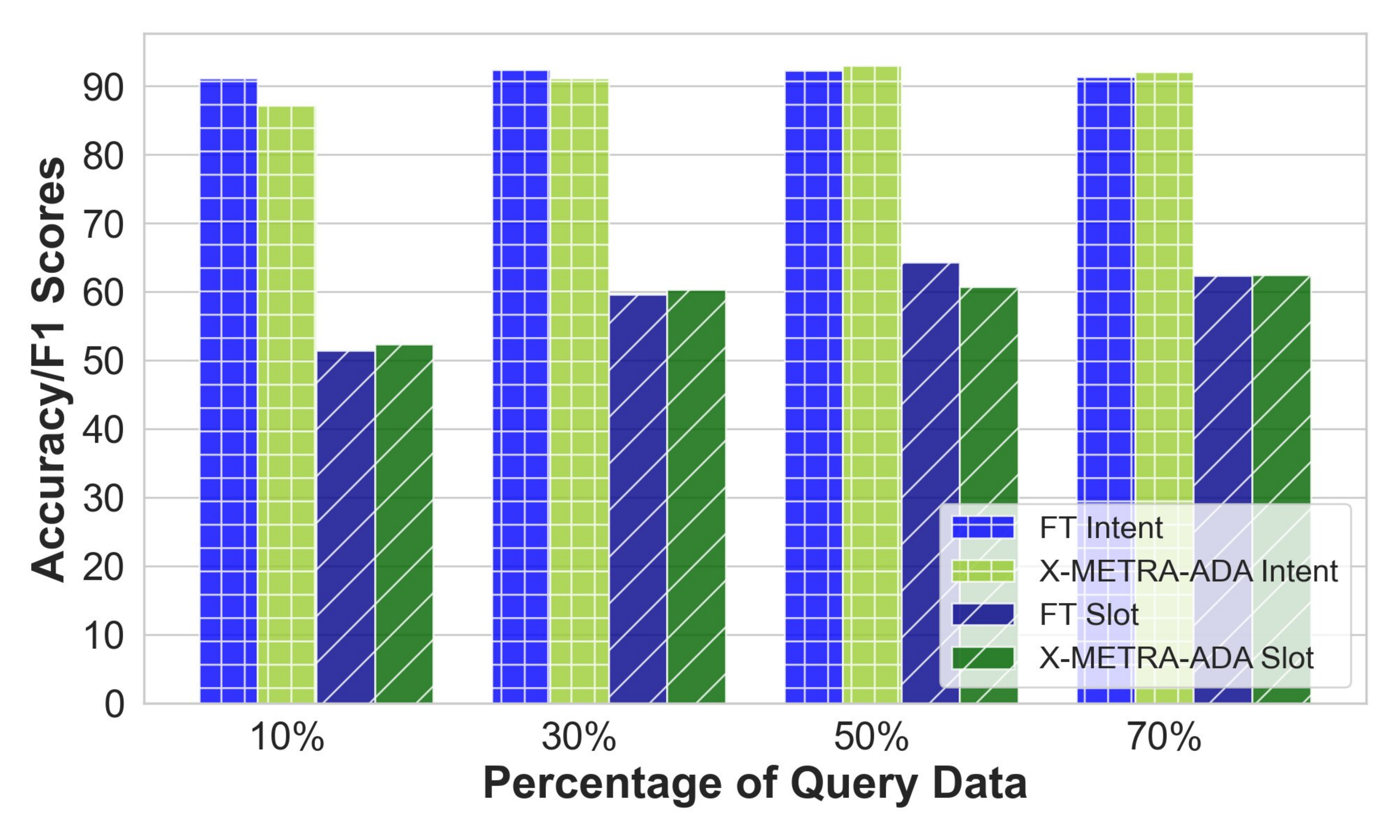}
 \caption{Downsampling Analysis for Few-shot on Spanish with Different Percentages of Query data.}
 \label{fig:down-es}
\end{figure}

Figure \ref{fig:down-es} shows a downsampling analysis on Spanish. Due to the typological similarity between Spanish and English, even lower percentages starting from $50\%$ of the query reach a maximal performance for both intent classification and slot filling.    

\paragraph{BERTology Analysis}
\begin{figure}
\centering
\includegraphics[width=0.5\textwidth]{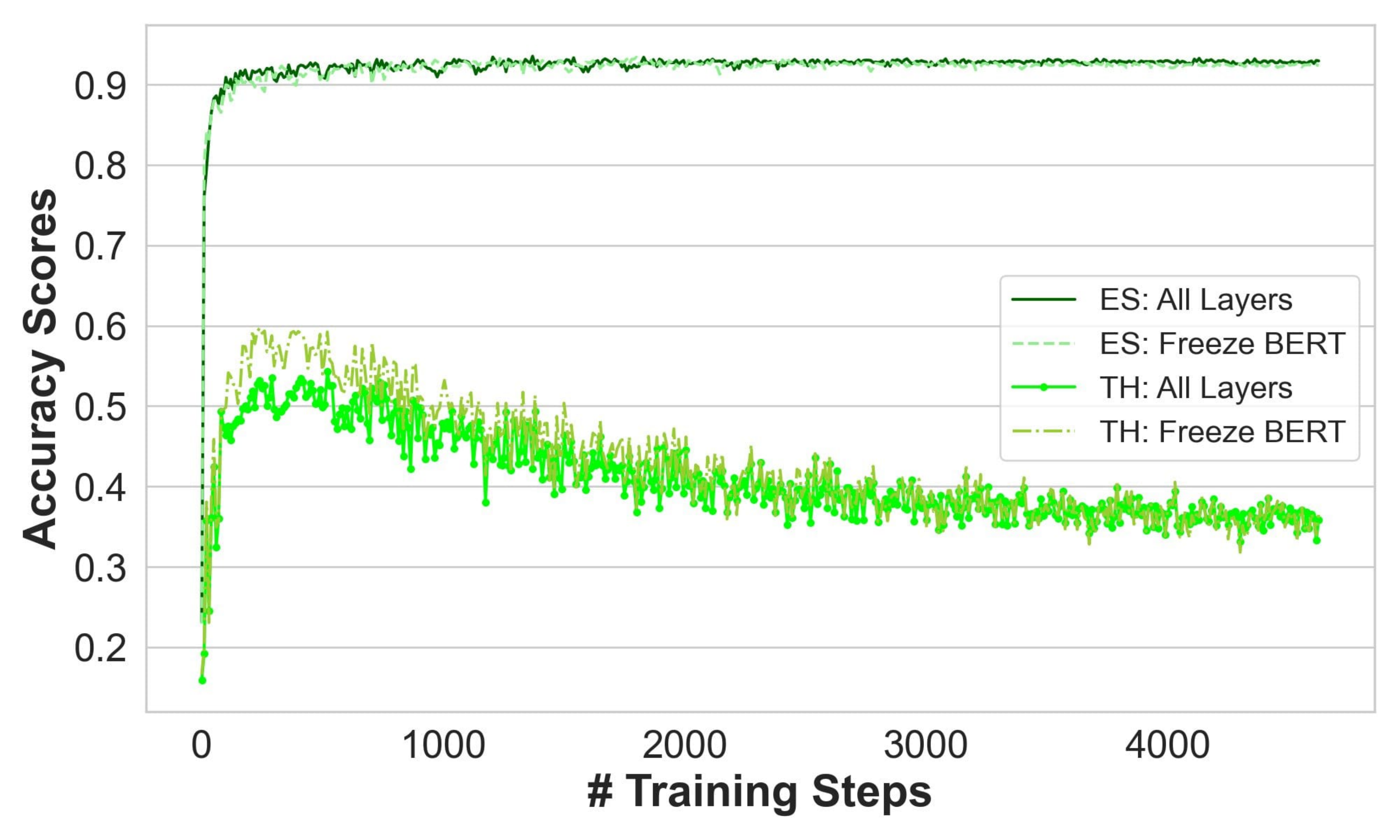}
 \caption{\label{fig:freeze-bert}The effect of freezing BERT layers of X-METRA-ADA during few-shot on intent classification.}
\end{figure}

We analyze the degree of contribution of M-BERT layers by freezing each pair of layers separately. Our analysis is not conclusive as the performance doesn't change significantly between layers. We then proceed to freeze all layers of M-BERT to discover that linear layers are more important in refining the cross-lingual alignment to the target language as shown by the narrow gap between freezing vs non-freezing BERT layers in Figure~\ref{fig:freeze-bert}. 
This can be explained by the challenge of fine-tuning M-BERT alone with many layers and higher dimensionality for such a low-resource setting.

\end{document}


\appendix
\section{Results on in-House Intent Classification Dataset}
\label{app-inhouse}
We perform an extensive evaluation including other languages for intent classification. We use an in-house dataset covering 7 languages including English with statistics of train/dev/test splits shown in Table ~\ref{inhouse-data}. \ref{in-house-results-meta} shows better performance in the favor of X-METRA-ADA with an average cross-lingual gain of $13.5$ in accuracy over the pre-training baseline on English only. Evaluation on more languages for task-oriented dialog shows some complicity trends between languages from the same family in addition to the already observed trend. As with the previous dataset, where the few-shot on the language of interest leads to the best performance, we notice higher numbers on the diagonal in the confusion matrix. In addition to that, we notice that languages like Japanese and Korean help each other where few-shot on one helps zero-shot on the other by a margin of 15.6 and 5.6 on Korean and Japanese respectively. 

\begin{table}[h]
\centering
\small
\begin{tabular}{llll}
\toprule
\textbf{Lang} & \textbf{Train} & \textbf{Dev} & \textbf{Test} \\ \toprule
English       & 5,438          & 1,814        & 1,814         \\
German        & 1,570          & 526          & 526           \\
French        & 1,082          & 362          & 362           \\
Italian       & 1,082          & 362          & 362           \\
Portuguese    & 1,150          & 386          & 386           \\
Japanese      & 1,070          & 358          & 358           \\
Korean        & 938            & 314          & 314           \\ \bottomrule
\end{tabular}
\caption{Statistics of In-House multilingual intent classification Dataset}
\label{inhouse-data}
\end{table}

\begin{table*}[]
\centering
\begin{tabular}{l|l|l|l|l|l|l|l}  \toprule
\multirow{2}{*}{\textbf{Type}}          & \multirow{2}{*}{\textbf{Model}}   &\multicolumn{6}{l} {\textbf{Test on}}
\\ \cline{3-8} 
                                                      &       &  DE & FR & IT & PT & JA & KR \\ \toprule
\multicolumn{1}{l|}{PRE} & EN & 19.1       & 30.0       & 30.1       & 26.1       & 14.6       & 5.1        \\ \hline
\multicolumn{1}{l|}{\multirow{6}{*}{X-METRA-ADA}}     & DE     & \textbf{34.3}       & 33.3       & 30.0       & 30.2       & 13.5       & 8.9        \\ \cline{2-8} 
\multicolumn{1}{l|}{}                          & FR     & 19.2       & \textbf{34.1}       & 29.9       & 29.1       & 5.8        & 9.0        \\ \cline{2-8} 
\multicolumn{1}{l|}{}                          & IT     & 18.3       & 32.2       & \textbf{44.4}       & 30.2       & 6.7        & 10.2       \\ \cline{2-8} 
\multicolumn{1}{l|}{}                          & PT     & 19.1       & 27.7       & 30.1       & \textbf{31.4}       & 5.8        & 9.0        \\ \cline{2-8} 
\multicolumn{1}{l|}{}                          & JA     & 24.1       & 25.7       & 33.2       & 26.1       & \textbf{30.9}       & 20.7       \\ \cline{2-8} 
\multicolumn{1}{l|}{}                          & KR     & 24.4       & 25.6       & 34.4       & 25.0       & 20.2       & \textbf{30.7}       \\ \hline

\multicolumn{1}{l|}{\multirow{6}{*}{ProtoNets}}     & DE     &   \underline{30.1}    &  23.4    &    21.9   &   \underline{35.5}    &  10.2      &  9.0     \\ \cline{2-8} 
\multicolumn{1}{l|}{}                          & FR     &  23.0     &  \underline{30.1}      &   21.2     &   32.1   &  15.5      &   9.1  \\ \cline{2-8} 
\multicolumn{1}{l|}{}                          & IT     &   27.5     &    26.7    &  \underline{22.1}     &     33.1 & 14.5     &   10.2       \\ \cline{2-8} 
\multicolumn{1}{l|}{}                          & PT     &  25.1    & 22.0       &  21.9     &    33.5    & 13.5        &   12.8      \\ \cline{2-8} 
\multicolumn{1}{l|}{}                          & JA     & 25.4       &  23.3       &  21.1      &   29.2     &      \underline{20.1}  &     10.2   \\ \cline{2-8} 
\multicolumn{1}{l|}{}                          & KR     &  25.1      &  22.1      & 20.3       &     26.0   &   14.5     & \underline{12.2} \\
\bottomrule
\end{tabular}
\caption{ Performance Evaluation on an In-House multilingual intent data between different meta-learning approaches MAML and prototypical. Bold results highlight best results for each test language whereas underlined results highlight second best results on ProtoNets~\cite{snell-prototypical-2017}.\label{in-house-results-meta}}
\end{table*}

\section{Full results for QA}
\label{app-qa}

Tables~\ref{qa-meta-f1-full} and \ref{qa-meta-em-full} shows the full results for F1 and Exact Match (EM) metrics for QA respectively.

\begin{table*}[h!] 
\small
\centering

\begin{tabular}{l|l|lllllll|l} \toprule
\multirow{2}{*}{\textbf{Type}}         & \multirow{2}{*}{\textbf{Model}}  &\multicolumn{7}{c|} {\textbf{Test on}} & \multirow{2}{*}{\textbf{Average}}
\\ \cline{3-9}
&  & AR & BN & FI & ID  &  RU & SW & TE &  \\ \toprule \multicolumn{1}{l|}{\multirow{2}{*}{EXT}}  & M-BERT                                              & 62.2  & 49.3 & 59.7 & 64.8   & 60.0 & 57.5  & 49.6 & 57.6  \\ 
&  MMTE & 63.1 & 55.8 & 53.9  & 60.9 & 58.9 & 63.1 & 54.2 & 58.6 \\ \hline
 PRE    & EN                                                      & 62.4 $\pm 2.2$ & 32.9 $\pm 1.4$ & 57.7 $\pm 4.4$ & 67.8 $\pm 3.8$ & 58.2 $\pm 3.7$ & 55.5 $\pm 2.9$ & 33.0 $\pm 5.9$ & 52.5 $\pm 3.5$ \\ \hline   
 \multirow{7}{*}{MONO}  & AR                                 &   74.0 $ \pm 1.1 $ &
30.1 $ \pm 2.4 $ &
50.0 $ \pm 0.8 $ &
59.5 $ \pm 1.3 $ &
48.4 $ \pm 0.8 $ &
50.8 $ \pm 1.7 $ &
24.1 $ \pm 2.7 $ &
48.1 $ \pm 1.5 $
 \\ 
\multicolumn{1}{l|}{}   & BN                                                                                           & 32.2 $ \pm 2.6 $ &
38.9 $ \pm 0.8 $ &
33.9 $ \pm 1.4 $ &
36.3 $ \pm 1.5 $ &
31.8 $ \pm 1.4 $ &
37.2 $ \pm 1.8 $ &
34.7 $ \pm 4.2 $ &
35.0 $ \pm 2.0 $
      \\ 
\multicolumn{1}{l|}{}   & FI                                                                                           & 54.2 $ \pm 2.5 $ &
30.7 $ \pm 1.3 $ &
63.3 $ \pm 1.5 $ &
52.5 $ \pm 1.7 $ &
43.0 $ \pm 2.1 $ &
48.6 $ \pm 1.7 $ &
28.7 $ \pm 2.8 $ &
45.9 $ \pm 1.9 $
    \\ 
\multicolumn{1}{l|}{}   & ID                                                                                         &   58.0 $ \pm 1.8 $ &
31.8 $ \pm 0.5 $ &
48.2 $ \pm 2.0 $ &
67.1 $ \pm 1.9 $ &
45.1 $ \pm 1.8 $ &
50.3 $ \pm 1.8 $ &
29.4 $ \pm 2.7 $ &
47.1 $ \pm 1.8 $
  \\ 
\multicolumn{1}{l|}{}   & RU                                                                                          &  50.9 $ \pm 2.3 $ &
34.5 $ \pm 2.1 $ &
45.2 $ \pm 4.2 $ &
52.0 $ \pm 4.0 $ &
54.4 $ \pm 1.3 $ &
47.1 $ \pm 2.1 $ &
30.7 $ \pm 2.5 $ &
45.0 $ \pm 2.6 $
  \\ 
\multicolumn{1}{l|}{}   & SW                                                                                            & 35.8 $ \pm 1.5 $ &
27.6 $ \pm 1.5 $ &
33.6 $ \pm 2.1 $ &
37.4 $ \pm 1.9 $ &
25.7 $ \pm 1.7 $ &
60.3 $ \pm 1.2 $ &
13.2 $ \pm 2.3 $ &
33.4 $ \pm 1.7 $
   \\ 
\multicolumn{1}{l|}{}   & TE                                                                                            &34.0 $ \pm 0.9 $ &
38.0 $ \pm 2.2 $ &
39.5 $ \pm 0.6 $ &
35.3 $ \pm 1.1 $ &
35.9 $ \pm 1.1 $ &
43.5 $ \pm 1.0 $ &
61.4 $ \pm 1.0 $ &
41.1 $ \pm 1.1 $
 \\ \hline   
 \multirow{7}{*}{FT}  & AR                                 &   \underline{77.0} $ \pm 0.3 $ &
36.8 $ \pm 2.9 $ &
58.8 $ \pm 0.6 $ &
67.0 $ \pm 2.7 $ &
60.9 $ \pm 0.8 $ &
52.4 $ \pm 3.6 $ &
32.0 $ \pm 1.0 $ &
55.0 $ \pm 1.7 $
\\ 
\multicolumn{1}{l|}{}   & BN                                                                                           &  60.7 $ \pm 0.4 $ &
\underline{51.0} $ \pm 2.7 $ &
59.2 $ \pm 0.6 $ &
67.1 $ \pm 1.6 $ &
59.2 $ \pm 0.3 $ &
56.2 $ \pm 0.8 $ &
43.7 $ \pm 0.9 $ &
56.7 $ \pm 1.0 $ \\ 
\multicolumn{1}{l|}{}   & FI                                                                                           &   60.3 $ \pm 1.9 $ &
36.7 $ \pm 1.3 $ &
\underline{70.9} $ \pm 0.4 $ &
65.7 $ \pm 1.4 $ &
62.1 $ \pm 0.5 $ &
50.9 $ \pm 1.3 $ &
36.4 $ \pm 3.6 $ &
54.7 $ \pm 1.5 $
       \\ 
\multicolumn{1}{l|}{}   & ID                                                                                            &  65.7 $ \pm 1.4 $ &
37.0 $ \pm 1.1 $ &
60.8 $ \pm 0.2 $ &
\underline{77.0} $ \pm 0.4 $ &
61.1 $ \pm 0.5 $ &
56.8 $ \pm 1.0 $ &
36.7 $ \pm 0.4 $ &
56.4 $ \pm 0.7 $
  \\ 
\multicolumn{1}{l|}{}   & RU                                                                                            &   60.9 $ \pm 2.5 $ &
37.2 $ \pm 2.0 $ &
59.0 $ \pm 2.1 $ &
66.8 $ \pm 1.3 $ &
\underline{64.8} $ \pm 0.4 $ &
55.2 $ \pm 1.8 $ &
36.8 $ \pm 1.3 $ &
54.4 $ \pm 1.6 $
  \\ 
\multicolumn{1}{l|}{}   & SW                                                                                            & 57.4 $ \pm 0.5 $ &
35.2 $ \pm 1.5 $ &
56.2 $ \pm 1.0 $ &
65.4 $ \pm 1.8 $ &
58.8 $ \pm 0.8 $ &
\underline{70.2} $ \pm 1.7 $ &
33.1 $ \pm 2.8 $ &
53.8 $ \pm 1.4 $
    \\ 
\multicolumn{1}{l|}{}   & TE                                                                                            &    54.0 $ \pm 3.2 $ &
39.1 $ \pm 2.1 $ &
54.8 $ \pm 2.3 $ &
63.5 $ \pm 2.6 $ &
58.1 $ \pm 0.9 $ &
56.9 $ \pm 1.8 $ &
\underline{65.4} $ \pm 0.6 $ &
56.0 $ \pm 1.9 $
 \\ \midrule \hline
\multicolumn{1}{l|}{\multirow{8}{*}{X-METRA-ADA}}     & AR                                                                                        & \textbf{78.4} $ \pm 0.6 $ &
33.0 $ \pm 0.8 $ &
58.2 $ \pm 0.2 $ &
66.4 $ \pm 1.4 $ &
59.9 $ \pm 0.1 $ &
53.2 $ \pm 3.8 $ &
31.4 $ \pm 3.0 $ &
54.4 $ \pm 1.4 $
 \\ 
\multicolumn{1}{l|}{}   & BN                                                                                         & 56.9 $ \pm 3.2 $ &
\textbf{53.2} $ \pm 0.5 $ &
56.7 $ \pm 1.4 $ &
67.4 $ \pm 1.2 $ &
56.7 $ \pm 1.3 $ &
56.0 $ \pm 0.9 $ &
41.7 $ \pm 0.6 $ &
55.5 $ \pm 1.3 $
 \\ 
\multicolumn{1}{l|}{}   & FI                                                                                           &    58.9 $ \pm 0.6 $ &
33.6 $ \pm 1.1 $ &
\textbf{72.8} $ \pm 0.3 $ &
61.9 $ \pm 2.0 $ &
60.7 $ \pm 0.9 $ &
46.5 $ \pm 1.2 $ &
36.6 $ \pm 1.7 $ &
53.0 $ \pm 1.1 $
  \\ 
\multicolumn{1}{l|}{}   & ID                                                                                          &     65.8 $ \pm 0.3 $ &
35.0 $ \pm 2.2 $ &
60.5 $ \pm 0.9 $ &
\textbf{77.7} $ \pm 0.2 $ &
60.4 $ \pm 1.3 $ &
57.4 $ \pm 1.1 $ &
35.3 $ \pm 0.3 $ &
56.0 $ \pm 0.9 $    
\\ 
\multicolumn{1}{l|}{}   & RU                                                                                          &  60.3 $ \pm 1.6 $ &
37.2 $ \pm 0.7 $ &
59.1 $ \pm 0.3 $ &
66.8 $ \pm 0.8 $ &
\textbf{66.2} $ \pm 0.1 $ &
53.7 $ \pm 0.8 $ &
33.2 $ \pm 3.1 $ &
53.8 $ \pm 1.1 $
 \\ 
\multicolumn{1}{l|}{}   & SW                                                                                         &  58.5 $ \pm 0.0 $ &
36.9 $ \pm 1.2 $ &
56.0 $ \pm 0.2 $ &
64.8 $ \pm 0.7 $ &
58.4 $ \pm 0.4 $ &
\textbf{71.9} $ \pm 0.2 $ &
33.7 $ \pm 1.5 $ &
54.3 $ \pm 0.6 $ \\

\multicolumn{1}{l|}{}   & TE                                                                                           & 56.0 $ \pm 3.0 $ &
38.8 $ \pm 0.1 $ &
53.6 $ \pm 1.7 $ &
61.1 $ \pm 1.9 $ &
58.6 $ \pm 0.6 $ &
55.8 $ \pm 0.2 $ &
\textbf{66.4} $ \pm 0.5 $ &
55.8 $ \pm 1.1 $
 \\ 
\bottomrule
\end{tabular}
\caption{\label{qa-meta-f1-full} F1 comparison on TyDiQA between external, pre-training, monolingual and fine-tuning baselines on one hand and X-METRA-ADA on the other hand.}
\end{table*}

\begin{table*}[h!] 
\small
\centering

\begin{tabular}{l|l|lllllll|l} \toprule
\multirow{2}{*}{\textbf{Type}}         & \multirow{2}{*}{\textbf{Model}}  &\multicolumn{7}{c|} {\textbf{Test on}} & \multirow{2}{*}{\textbf{Average}}
\\ \cline{3-9}
&  & AR & BN & FI & ID  &  RU & SW & TE &  \\ \toprule \multicolumn{1}{l|}{\multirow{2}{*}{EXT}}  & M-BERT                      &  42.8   & 32.7 &  45.3 &  45.8  & 38.8 & 37.9  & 38.4  & 40.2                   \\ 
&  MMTE  &   39.2 &  41.9  & 42.1 &  47.6  &  37.9 &  47.2 &  45.8 &  43.1 \\ \hline
 PRE    & EN                                                      & 43.0 $ \pm 0.4 $ &
16.8 $ \pm 1.8 $ &
39.6 $ \pm 1.2 $ &
49.2 $ \pm 1.1 $ &
35.5 $ \pm 0.9 $ &
35.1 $ \pm 1.0 $ &
18.6 $ \pm 0.5 $ &
34.0 $ \pm 1.0 $
 \\ \hline   
 \multirow{7}{*}{MONO}  & AR                                 & 57.5 $ \pm 1.5 $ &
19.7 $ \pm 2.9 $ &
35.1 $ \pm 1.0 $ &
44.2 $ \pm 1.3 $ &
25.2 $ \pm 0.9 $ &
33.8 $ \pm 1.4 $ &
14.9 $ \pm 1.7 $ &
32.9 $ \pm 1.5 $

 \\ 
\multicolumn{1}{l|}{}   & BN                                                                                           & 17.1 $ \pm 1.4 $ &
24.5 $ \pm 2.9 $ &
17.5 $ \pm 0.4 $ &
20.8 $ \pm 2.0 $ &
14.4 $ \pm 0.5 $ &
20.5 $ \pm 1.4 $ &
19.9 $ \pm 5.0 $ &
19.2 $ \pm 1.9 $

      \\ 
\multicolumn{1}{l|}{}   & FI                                                                                          & 33.7 $ \pm 4.0 $ &
15.6 $ \pm 1.6 $ &
49.8 $ \pm 1.3 $ &
35.3 $ \pm 2.3 $ &
21.4 $ \pm 1.4 $ &
26.1 $ \pm 9.9 $ &
16.5 $ \pm 3.9 $ &
28.3 $ \pm 3.5 $

    \\ 
\multicolumn{1}{l|}{}   & ID                                                                                         & 39.7 $ \pm 1.4 $ &
18.6 $ \pm 1.3 $ &
32.7 $ \pm 1.9 $ &
54.9 $ \pm 0.1 $ &
23.8 $ \pm 0.6 $ &
34.4 $ \pm 1.2 $ &
16.9 $ \pm 4.9 $ &
31.6 $ \pm 1.6 $

  \\ 
\multicolumn{1}{l|}{}   & RU                                                                                          & 30.8 $ \pm 1.9 $ &
26.3 $ \pm 4.9 $ &
29.7 $ \pm 2.4 $ &
34.9 $ \pm 4.0 $ &
37.9 $ \pm 1.6 $ &
30.7 $ \pm 3.1 $ &
19.9 $ \pm 1.9 $ &
30.0 $ \pm 2.8 $

  \\ 
\multicolumn{1}{l|}{}   & SW                                                                                            & 16.0 $ \pm 1.3 $ &
16.5 $ \pm 1.5 $ &
15.6 $ \pm 1.0 $ &
21.1 $ \pm 1.3 $ &
10.5 $ \pm 1.3 $ &
48.6 $ \pm 1.2 $ &
5.3 $ \pm 1.7 $ &
19.1 $ \pm 1.3 $

   \\ 
\multicolumn{1}{l|}{}   & TE                                                                                            & 18.8 $ \pm 2.0 $ &
26.3 $ \pm 1.5 $ &
23.8 $ \pm 2.6 $ &
21.6 $ \pm 2.5 $ &
20.4 $ \pm 1.2 $ &
26.7 $ \pm 1.7 $ &
46.3 $ \pm 1.1 $ &
26.3 $ \pm 1.8 $
 \\ \hline   
 \multirow{7}{*}{FT}  & AR                                 & \underline{61.3} $ \pm 1.0 $ &
26.5 $ \pm 4.4 $ &
43.1 $ \pm 1.0 $ &
52.2 $ \pm 2.0 $ &
37.9 $ \pm 2.5 $ &
35.6 $ \pm 3.3 $ &
21.0 $ \pm 3.0 $ &
39.7 $ \pm 2.5 $

\\ 
\multicolumn{1}{l|}{}   & BN                                                                                          & 42.2 $ \pm 0.9 $ &
\underline{38.0} $ \pm 4.4 $ &
44.8 $ \pm 1.2 $ &
51.5 $ \pm 2.2 $ &
36.8 $ \pm 1.6 $ &
37.2 $ \pm 1.7 $ &
27.3 $ \pm 0.2 $ &
39.7 $ \pm 1.7 $
 \\ 
\multicolumn{1}{l|}{}   & FI                                                                                          & 43.2 $ \pm 1.8 $ &
23.6 $ \pm 1.1 $ &
\underline{56.5} $ \pm 0.6 $ &
50.8 $ \pm 2.1 $ &
40.5 $ \pm 0.8 $ &
33.5 $ \pm 1.2 $ &
20.7 $ \pm 3.3 $ &
38.4 $ \pm 1.6 $

       \\ 
\multicolumn{1}{l|}{}   & ID                                                                                            & 49.4 $ \pm 1.6 $ &
23.3 $ \pm 2.4 $ &
46.4 $ \pm 0.4 $ &
\underline{63.8} $ \pm 0.5 $ &
40.5 $ \pm 0.1 $ &
38.1 $ \pm 2.1 $ &
24.1 $ \pm 0.5 $ &
40.8 $ \pm 1.1 $

  \\ 
\multicolumn{1}{l|}{}   & RU                                                                                            & 42.6 $ \pm 2.6 $ &
24.8 $ \pm 3.3 $ &
43.5 $ \pm 2.0 $ &
52.4 $ \pm 2.3 $ &
\underline{46.5} $ \pm 0.4 $ &
37.6 $ \pm 1.5 $ &
24.5 $ \pm 1.3 $ &
38.8 $ \pm 1.9 $

  \\ 
\multicolumn{1}{l|}{}   & SW                                                                                            & 38.9 $ \pm 0.6 $ &
23.0 $ \pm 1.4 $ &
40.1 $ \pm 1.4 $ &
50.0 $ \pm 1.7 $ &
38.0 $ \pm 0.8 $ &
\underline{59.0} $ \pm 3.1 $ &
23.5 $ \pm 1.4 $ &
38.9 $ \pm 1.5 $

    \\ 
\multicolumn{1}{l|}{}   & TE                                                                                            & 36.1 $ \pm 2.2 $ &
30.0 $ \pm 2.3 $ &
40.0 $ \pm 2.5 $ &
49.4 $ \pm 2.1 $ &
38.6 $ \pm 0.9 $ &
39.0 $ \pm 1.7 $ &
\underline{49.2} $ \pm 0.5 $ &
40.3 $ \pm 1.7 $
 
 \\ \midrule \hline
\multicolumn{1}{l|}{\multirow{8}{*}{X-METRA-ADA}}     & AR                                                                                       & \textbf{63.3} $ \pm 0.8 $ &
21.2 $ \pm 1.9 $ &
42.6 $ \pm 1.0 $ &
51.8 $ \pm 1.2 $ &
34.9 $ \pm 1.1 $ &
36.0 $ \pm 3.5 $ &
20.9 $ \pm 1.7 $ &
38.7 $ \pm 1.6 $
 \\
\multicolumn{1}{l|}{}   & BN                                                                                         & 29.2 $ \pm 16.5 $ &
\textbf{39.0} $ \pm 1.9 $ &
41.9 $ \pm 1.6 $ &
51.1 $ \pm 1.7 $ &
34.1 $ \pm 0.4 $ &
37.1 $ \pm 1.4 $ &
25.6 $ \pm 0.2 $ &
36.9 $ \pm 3.4 $
 \\
\multicolumn{1}{l|}{}   & FI                                                                                           & 42.0 $ \pm 1.0 $ &
20.4 $ \pm 0.7 $ &
\textbf{59.1} $ \pm 1.1 $ &
46.0 $ \pm 2.7 $ &
36.8 $ \pm 1.3 $ &
30.9 $ \pm 0.6 $ &
22.5 $ \pm 0.9 $ &
36.3 $ \pm 1.8 $
 \\
\multicolumn{1}{l|}{}   & ID                                                                                          & 54.8 $ \pm 7.9 $ &
20.1 $ \pm 1.5 $ &
46.1 $ \pm 1.2 $ &
\textbf{65.2} $ \pm 0.5 $ &
38.5 $ \pm 1.9 $ &
39.6 $ \pm 0.8 $ &
23.1 $ \pm 1.4 $ &
41.1 $ \pm 2.2 $
   \\ 
\multicolumn{1}{l|}{}   & RU                                                                                         & 42.9 $ \pm 1.3 $ &
26.5 $ \pm 1.2 $ &
43.0 $ \pm 0.6 $ &
53.0 $ \pm 0.1 $ &
\textbf{48.9} $ \pm 0.4 $ &
35.3 $ \pm 1.0 $ &
21.6 $ \pm 2.4 $ &
38.7 $ \pm 1.0 $
 \\
\multicolumn{1}{l|}{}   & SW                                                                                        &39.9 $ \pm 0.4 $ &
26.0 $ \pm 1.1 $ &
40.0 $ \pm 0.7 $ &
50.3 $ \pm 0.4 $ &
38.0 $ \pm 0.9 $ &
\textbf{61.4} $ \pm 0.4 $ &
23.9 $ \pm 0.7 $ &
39.9 $ \pm 0.7 $
 \\ 
\multicolumn{1}{l|}{}   & TE                                                                                         & 38.0 $ \pm 3.9 $ &
28.3 $ \pm 0.0 $ &
37.0 $ \pm 2.3 $ &
47.6 $ \pm 3.4 $ &
36.3 $ \pm 0.5 $ &
36.9 $ \pm 1.2 $ &
\textbf{49.7} $ \pm 0.5 $ &
39.1 $ \pm 1.7 $
  \\ 
\bottomrule
\end{tabular}
\caption{\label{qa-meta-em-full} EM comparison on TyDiQA between external, pre-training, monolingual and fine-tuning baselines on one hand and X-METRA-ADA on the other hand.}
\end{table*}

\begin{table*}[t!]
\vspace{-0.4cm}
\centering
\scalebox{0.8}{
\begin{tabular}{l|l|ll|ll|ll}  \toprule
\multirow{2}{*}{\textbf{Type}}          & \multirow{2}{*}{\textbf{Model}}                                                                                         & \multicolumn{2}{c|}{\textbf{Test on ES}} & \multicolumn{2}{c|}{\textbf{Test on TH}} & \multicolumn{2}{c}{\textbf{Average}} \\
                           &                                                                                                           & Intent Acc      & Slot F1      & Intent Acc      & Slot F1 & Intent Acc      & Slot F1      \\ \toprule 
\multicolumn{1}{c}{Zero-shot Learning}   \\ \midrule                 
\multicolumn{1}{l|}{\multirow{3}{*}{EXT}} & MCoVe   & 53.9            & 19.3         & 70.7            & 35.6   & 62.3 & 27.5      \\ 
 \multicolumn{1}{l|}{}                          & MLT_{H}* & 82.9            & \textbf{74.9}         & 53.8            & 26.1  & 68.4   & 50.5   \\  
  \multicolumn{1}{l|}{}                          & MLT_{A}*  & \underline{87.9}            & 73.9         &  \underline{73.5}            & 27.1 &  80.7 & \underline{50.5} \\ \multicolumn{1}{l|}{}                          & MMTE &  93.6             &    -       &  89.6        &  -  &  \underline{91.6}   &  -   \\ \midrule
\multicolumn{1}{l|}{\multirow{1}{*}{PRE}}   &  EN                                                                                        & 66.3            & 31.9         & 39.8            & 12.1  & 53.0 &  22.0    \\ \midrule

\multicolumn{1}{c}{Few-shot Learning}  \\ \midrule

\multicolumn{1}{l|}{\multirow{2}{*}{MONO}}   & ES                                                                                    &  82.4 $\pm 6.0$          & 43.9 $\pm 1.5$       &  38.7 $\pm 15.4$            & 9.6 $\pm 1.7$ & \multicolumn{1}{l}{\multirow{2}{*}{80.1 $\pm 5.3$}} & \multicolumn{1}{l}{\multirow{2}{*}{46.3 $\pm 2.7$}}
  \\ 
  
  \multicolumn{1}{l|}{}   & TH                                                                                     &   39.7 $\pm 7.9$          &  5.9 $\pm 1.1$      &   79.1 $\pm 4.7 $         & \underline{54.1} $\pm 3.9$ &  & 
  \\ \midrule
\multicolumn{1}{l|}{\multirow{2}{*}{FT OLD}}    & ES                                                                                      &  74.0  $\pm 1.1$          &   41.3  $\pm 0.1$        &  34.1  $\pm 1.1$           &  15.3 $\pm 0.2$   & \multicolumn{1}{l}{\multirow{2}{*}{67.8 $\pm 0.6$}} & \multicolumn{1}{l}{\multirow{2}{*}{38.8 $\pm 0.2$}}      \\ 
  
  \multicolumn{1}{l|}{}   & TH                                                                                      & 71.0 $\pm 1.0$            &   31.4 $\pm 0.1$        &  61.7   $\pm 0.1$           &   36.3 $\pm 0.2$  &  &  
  \\  \midrule 
  \multicolumn{1}{l|}{\multirow{3}{*}{FT NEW}}    & ES                                                                                      &  89.7          &  56.8         &   52.5      &  11.8   & \multicolumn{1}{l}{\multirow{2}{*}{86.6}} & \multicolumn{1}{l}{\multirow{2}{*}{57.9}}      \\ 
  
  \multicolumn{1}{l|}{}   & TH                                                                                      &      42.8      &   24.8       &      83.5        &  59.0   &   &   
 \\ 
  
  \multicolumn{1}{l|}{}   & JOINT                                                                                      &   85.0         &  51.8      &     78.3          &  53.1   &  81.7 &  52.5 
  \\  \midrule 
\multicolumn{1}{l|}{\multirow{4}{*}{X-METRA-ADA}}    & ES                                                                                            &   \textbf{91.9}  $\pm 0.4$            & \underline{55.9}  $\pm 1.3$         & 56.2  $\pm 2.2$           & 12.2   $\pm 0.2$    & \multicolumn{1}{l}{\multirow{2}{*}{\textbf{90.9 $\pm 0.9$}}} & \multicolumn{1}{l}{\multirow{2}{*}{\textbf{58.6 $\pm 0.8$}}}      \\ 
\multicolumn{1}{l|}{}   & TH                                                                                            &  37.2  $\pm 4.4 $             &  30.7  $\pm 0.5 $        & \textbf{89.9}  $\pm 1.4 $         & \textbf{61.4}  $\pm 0.3 $       \\ \multicolumn{1}{l|}{}   & JOINT INTER & 87.1  & 55.9 &  87.7 & 58.7  & 87.4 &  57.3 \\ \multicolumn{1}{l|}{}   & JOINT ALL &  86.3 & 56.0 & 87.3  & 60.4  & 86.8 &  58.2 \\  \bottomrule
\end{tabular}
}
\caption{\label{facebook-results-meta} Performance evaluation on NLU between meta-learning approaches, fine-tuning internal baselines and external baselines. We highlight the best scores in bold and underline the second best for each language and sub-task. * is not entirely zero-shot as it uses mixed language training.}
\end{table*}

\section{Dataset Statistics}
\subsection{Multilingual Task-Oriented Dialogue}
Tables \ref{data-face} and \ref{data-mtop} show the statistics of MTOD and MTOP respectively per language and split. 

\begin{table}[H]
\small
\centering
\begin{tabular}{lllll}
\toprule
\textbf{Lang} & \textbf{ISO} & \textbf{Train} & \textbf{Dev} & \textbf{Test} \\ \toprule
English       & EN & 30,521         & 4,181        & 8,621         \\ 
Spanish       & ES & 3,617          & 1,983        & 3,043         \\
Thai          & TH & 2,156          & 1,235       & 1,692         \\ \bottomrule
\end{tabular}
\caption{Statistics of MTOD dataset~\cite{xlingualnlu-schuster-naacl19} per language and split.}
\label{data-face}
\end{table}

\begin{table}[H]
\small
\centering
\begin{tabular}{lllll}
\toprule
\textbf{Lang} & \textbf{ISO} & \textbf{Train} & \textbf{Dev} & \textbf{Test} \\ \toprule
English       & EN &     15,667  &   2,235      &   4,386       \\ 
German       & DE &  13,424        & 1,815       &  3,549       \\
French          & FR &  11,814        & 1,577       &  3,193        \\
Spanish          & ES & 10,934         &  1,527      &  2,998        \\
Hindi          & HI &   11,330       &  2,012      &  2,789        \\
Thai          & TH &  10,759        &  1,671      &   2,765       \\
\bottomrule
\end{tabular}
\caption{Statistics of MTOP dataset~\cite{li-mtop-20} per language and split.}
\label{data-mtop}
\end{table}

\subsection{Multilingual Question Answering}
Tables \ref{mlqa-data} and \ref{data-tydiqa} show the statistics of MLQA and TydiQA respectively per language and split. 

\begin{table}[H]
\small
\centering
\begin{tabular}{lllll}
\toprule
\textbf{Lang} & \textbf{ISO} & \textbf{Train} & \textbf{Dev} & \textbf{Test} \\ \toprule
English       & EN &          &  1,148       & 11,590        \\ 
Arabic       & AR &    -       &  517        &  5,335      \\
German      & DE &   -    &   512    &   4,517     \\ 
Spanish      &  ES &    -    &   500     &  5,253       \\ 
Hindi         & HI &   -      &  507      &  4,918       \\
Simplified Chinese & ZH &    -    &    504    &  5,137  \\ 
Vietnamese         & VI &   -    &   511     &   5,495     \\  \bottomrule
\end{tabular}
\caption{Statistics of MLQA dataset~\cite{lewis-mlqa-acl20} per language and split.}
\label{mlqa-data}
\end{table}

\begin{table}[H]
\small
\centering
\begin{tabular}{lllll}
\toprule
\textbf{Lang} & \textbf{ISO} & \textbf{Train} & \textbf{Dev} & \textbf{Test} \\ \toprule
English       & EN &   3,326       &  370       & 440        \\ 
Arabic       & AR &   13,324        & 1,481         & 921         \\
Bengali         & BN & 2,151      &  239     & 113       \\ 
Finnish         &  FI & 6,169       & 686      &  782      \\ 
Indonesian         & ID &  5,131        &  571     &  565       \\ 
Russian        & RU &  5,841       &  649     &  812      \\ 
Swahili        & SW &   2,479      &  276     &   499     \\ 
Telugu        & TE &   5,006      &  557     &   669     \\ \bottomrule
\end{tabular}
\caption{Statistics of TydiQA dataset per language and split.}
\label{data-tydiqa}
\end{table}
\bibliographystyle{acl_natbib}
\bibliography{bibfile1}